\documentclass{article}

\PassOptionsToPackage{numbers, compress}{natbib}

\usepackage[main,final]{neurips_2026}

\usepackage[utf8]{inputenc} 
\usepackage[T1]{fontenc}    
\usepackage{hyperref}       
\usepackage{url}            
\usepackage{booktabs}       
\usepackage{amsfonts}       
\usepackage{amssymb}        
\usepackage{amsmath}        
\usepackage{nicefrac}       
\usepackage{microtype}      
\usepackage[table]{xcolor}  
\usepackage{graphicx}       
\usepackage{multirow}       
\usepackage{float}
\usepackage{capt-of}

\usepackage{xspace}
\newcommand{\methodname}{RegimeVGGT\xspace}

\title{\methodname{}: Layer-Wise Spatially Preserving Redundancy Removal for Visual Geometry Grounded Transformer}

\author{%
  Jinhao You$^{1,*}$ \quad Shuo Lyu$^{1,*,\dagger}$ \quad
  Zhuohang Lyu$^{1,*}$ \quad Tanxuan Li$^{1,*}$ \\
  \bfseries Zibo Zhao$^{1,*}$ \quad Jiaxiang Hu$^{2}$ \quad
  Kai Tang$^{3}$ \quad Yichen Guo$^{3}$ \\[0.3em]
  {\normalfont $^{1}$University of Pennsylvania} \\
  {\normalfont $^{2}$University of California, Irvine} \\
  {\normalfont $^{3}$Nanyang Technological University}
}

\begin{document}

\maketitle

\begingroup
\renewcommand{\thefootnote}{\fnsymbol{footnote}}
\footnotetext[1]{Equal contribution.\quad
\textsuperscript{\textdagger} Corresponding author.}
\endgroup

\begin{abstract}
Visual Geometry Grounded Transformer (VGGT) recovers dense 3D scene structure from multi-view images in one forward pass, but quadratic cross-frame attention limits its scalability. Existing training-free accelerators exploit token, frame, or attention redundancy; we investigate how layer-wise functional roles can guide compression. Our spectral, probing, and causal analyses reveal three regimes: shallow layers lack cross-view structure, middle layers drive cross-view alignment, and deep layers are redundant for dense geometry yet their cross-frame attention remains essential for pose. RegimeVGGT applies layer-wise U-shaped compression along two axes: \emph{Saliency-Guided Banded Merging} protects geometry- and edge-salient tokens, while \emph{Selectively Protected K/V Downsampling} preserves cross-frame spatial coverage and the pose-critical path through a phase-shifted spatial grid, a reference-frame anchor, and uncompressed camera/register tokens. Without any training, RegimeVGGT achieves a $6.7\times$ speedup over the memory-optimized VGGT$^\ast$ at matched reconstruction quality. Project page: \url{https://tracyou07.github.io/regimevggt/}

\end{abstract}

\section{Introduction}
\label{sec:intro}

Recovering the 3D structure of a scene from multi-view images is fundamental to autonomous driving~\cite{geiger2013vision,caesar2020nuscenes}, robotics~\cite{thrun2002probabilistic,mur2017orb},
and augmented reality~\cite{izadi2011kinectfusion}, all of which demand efficient, end-to-end inference pipelines. Traditional multi-stage pipelines such as COLMAP~\cite{schonberger2016pixelwise,schonberger2016structure} involve feature matching, camera estimation, and dense reconstruction, with substantial computational demands at large scale~\cite{agarwal2011building}. Recent feed-forward models, pioneered by DUSt3R~\cite{wang2024dust3r} and extended by subsequent works~\cite{leroy2024grounding,wang2025continuous,yang2025fast3r}, regress dense 3D geometry directly from uncalibrated images using pairwise, parallel multi-view, or recurrent architectures.

VGGT~\cite{wang2025vggt} further unifies geometry and camera prediction in a transformer that accepts an arbitrary number of images and jointly predicts 3D attributes in a single forward pass. At its core is a 24-layer aggregator with alternating frame-wise and global cross-frame attention, where each patch token is geometrically grounded to a specific 3D surface point through dedicated prediction heads. This design incurs $O(S^2P^2)$ cost that grows quadratically with both the number of frames $S$ and patches per frame $P$. As a result, vanilla VGGT faces substantial memory demands on long sequences, while even the memory-optimized variant VGGT$^\ast$~\cite{shen2025fastvggt} remains prohibitively slow at this scale.

Concurrent training-free accelerators reduce VGGT's bottleneck along different axes: FastVGGT~\cite{shen2025fastvggt} merges tokens uniformly across all 24 layers, while S-VGGT~\cite{li2026s} partitions frames into independent subscenes. AVGGT~\cite{sun2025avggt} studies layer-wise redundancy in VGGT: it
  converts early global attention to per-frame mode and uniformly subsamples K/V on a spatial grid in the remaining layers. Yet two finer questions remain open: \emph{$\mathcal{Q}_1$: How does VGGT's aggregator partition along depth into functionally distinct regimes?} \emph{$\mathcal{Q}_2$: Within each such regime, how should its functional role and internal structure inform compression design?}

\begin{figure}[t]
  \centering
  \includegraphics[width=0.9\linewidth]{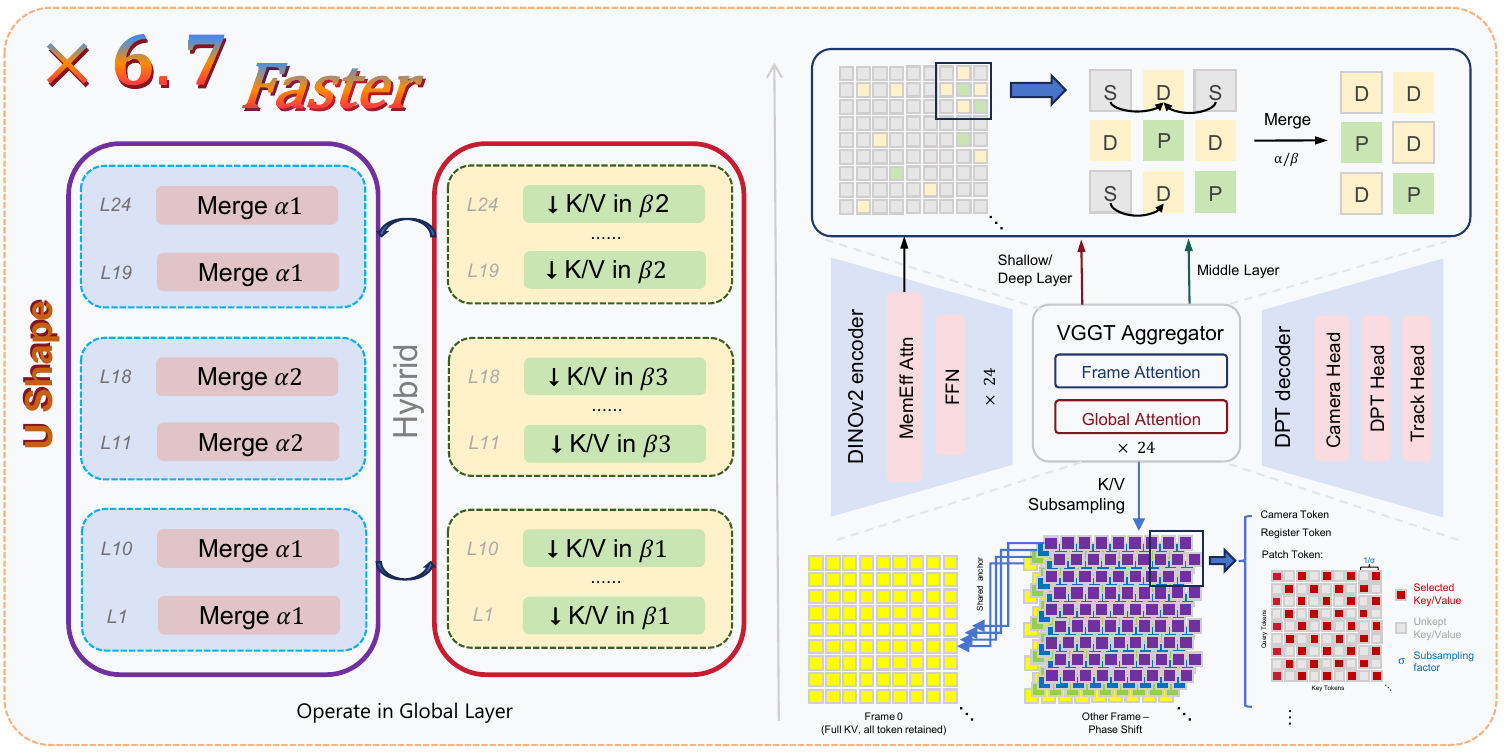}
\caption{\textbf{Overview of \methodname{}.} VGGT's 24 global-attention layers split into three bands: the shallow (L1--L10) and deep (L19--L24) bands have lower effective attention rank than the middle band (L11--L18), where cross-view alignment concentrates. This partition configures two training-free operators composed in every global block: \emph{Saliency-Guided Banded Merging} reduces the token count with a U-shaped per-band merge ratio and a band-aligned index cache, and \emph{Selectively Protected K/V Downsampling} sparsifies K/V with a phase-shifted per-frame grid, a full-density reference-frame anchor, and uncompressed camera/register tokens.}
  \label{fig:overview}
\end{figure}

We answer both through a spectral, probing, and causal analysis that uncovers three bands across the evaluated scenes. The middle band (L11--L18) drives cross-view geometric alignment, the shallow band (L1--L10) lacks strong cross-view structure, and the deep band (L19--L24) has lower sensitivity in the dense-output diagnostics but still requires cross-frame updates for camera-pose estimation. Each band exposes two complementary compressible axes (a \emph{token-count axis} and a \emph{K/V-density axis}) with different headroom: the middle band has high effective rank but concentrates information in a small set of geometry- and edge-salient tokens at 3D boundaries and along 2D image structure, while its dense attention limits K/V sparsification; the lower-rank shallow and deep bands admit stronger compression. Beyond these per-band patterns, pose stability imposes a constraint on compression: the cross-frame attention path between camera/register~\cite{darcet2023vision} tokens and the surrounding patches must remain available, even when the dense-output diagnostics indicate redundancy.

These findings lead to \methodname{}, a training-free framework with two operators composed within every global block and configured by the same three-band partition. \emph{Saliency-Guided Banded Merging} compresses the token-count axis with a U-shaped merge ratio that mirrors the rank profile (aggressive on the low-rank flanks), protects geometry- and edge-salient tokens via a frozen DINOv2 class-attention saliency $\Psi$ at zero aggregator-state cost, and amortizes matching by band-aligned caching. \emph{Selectively Protected K/V Downsampling} compresses the K/V-density axis under the same U-shaped schedule while preserving the pose path: queries always participate, but each non-anchor frame's patch K/V is restricted to a phase-shifted sub-grid that diversifies the sampled grid coordinates, while a full-density frame-$0$ anchor and uncompressed cam/reg tokens keep cam/reg--patch cross-frame attention intact. \methodname{} reaches a $6.7\times$ speedup over VGGT$^\ast$ on 1000-frame ScanNet-50.

\section{Related Work}
\label{sec:related}

\paragraph{Feed-forward 3D reconstruction.}
Classical multi-view reconstruction relies on feature matching, triangulation, and bundle adjustment~\cite{schonberger2016structure,schonberger2016pixelwise}. Learning-based MVS methods such as MVSNet~\cite{yao2018mvsnet} build cost volumes from calibrated views. Recent feed-forward approaches relax these constraints: DUSt3R~\cite{wang2024dust3r} predicts dense pointmaps from uncalibrated pairs, Fast3R~\cite{yang2025fast3r} and CUT3R~\cite{wang2025continuous} scale to hundreds of views, VGGT~\cite{wang2025vggt} jointly estimates geometry and camera poses, and $\pi^3$~\cite{wang2025pi} improves permutation robustness. For models using dense global cross-frame attention, the $O(S^2P^2)$ cost motivates acceleration that accounts for the token roles involved in 3D prediction.

\paragraph{Token reduction in vision transformers.}
Token reduction has been extensively explored for accelerating vision transformers. Pruning-based methods~\cite{rao2021dynamicvit} drop less informative tokens, reorganization approaches such as EViT~\cite{liang2022not} retain salient ones, and ToMe~\cite{tome} merges redundant tokens via training-free similarity. Later works extend token merging to generative models, pruning--merging hybrids, vision–language systems, and long-form video~\cite{bolya2023token,kim2024token,wang2025dymu,lee2024video}. Other studies examine token-reduction patterns across backbones and datasets~\cite{haurum2023tokens}, and token-level cross-modal fusion~\cite{wang2022multimodal}. These studies motivate accounting for token structure; in 3D reconstruction, our analysis examines the geometric distinction between boundary and smooth-region tokens.

\paragraph{Accelerating VGGT.}
Several concurrent works accelerate VGGT along different axes.
At the token level, FastVGGT~\cite{shen2025fastvggt} applies ToMe/ToMeSD-style bipartite merging~\cite{tome,bolya2023token} uniformly across layers, LiteVGGT~\cite{shu2025litevggt} combines geometry-aware cached merging with fine-tuning and FP8 inference, and HTTM~\cite{wang2025httm} tailors merging to VGGT's global attention via head-wise temporal merging, spatio-temporal reordering, and outlier filtering.
At the attention-pattern level, Faster VGGT~\cite{wang2025faster} replaces dense attention with SpargeAttention-style block-sparse kernels~\cite{zhang2025spargeattention} by exploiting geometry-concentrated probability mass.
At the frame level, S-VGGT~\cite{li2026s} partitions inputs into independently processed subscenes, while StreamVGGT~\cite{zhuo2025streaming} distills VGGT into a causal streaming aggregator.
At the layer level, AVGGT~\cite{sun2025avggt} converts early layers to frame attention but still relies on uniform grid-based K/V downsampling.
In contrast, our method simultaneously exploits heterogeneity along two compression axes.

\section{Method}
\label{sec:method}
\subsection{Analysis: The Case for Dual-Axis Compression}
\label{sec:analysis}

\paragraph{Layer-wise U-shape redundancy.}
We diagnose VGGT's 24 global-attention layers along three independent dimensions, all converging to a narrow middle band. \emph{Information availability} (ridge probe; Appendix~\ref{app:probe}) is recoverable mainly across L11--L18; \emph{information utilization} (key-permutation intervention; Appendix~\ref{app:causal}) peaks at the camera head there and is weaker on both flanks under single-layer perturbations; \emph{spectral structure} of the per-head attention probability matrices (Appendix~\ref{app:attention-rank}) is inverted-U-shaped, peaking in the middle and dropping sharply on both flanks. Cross-view geometric alignment is therefore concentrated in L11--L18; the flanks are redundant for \emph{distinct} reasons: shallow because cross-view content is not yet operative, deep because attention has collapsed into low-dimensional refinement. This U-shape redundancy invites two orthogonal compression axes, the \textbf{token axis} and the \textbf{K/V axis}.

\paragraph{Heterogeneous geometry tokens and middle-band concentration.}
Within the middle band not all patch tokens are equally important. We rank patches by the squared residual of regressing ground-truth depth on low-level 2D cues; the top subset, the \emph{geometry tokens}, concentrates around depth discontinuities, object contours, and high-frequency 3D boundaries not explainable by 2D cues (formal definition in Appendix~\ref{app:geo_token}). Figure~\ref{fig:geometry} verifies they (i) align spatially with depth edges rather than appearance texture, (ii) disrupt the depth head under ablation substantially more than both random ablation and a Sobel-ranked image-edge baseline on 3D-rich scenes, with the gap narrowing on geometrically flat scenes where 3D boundaries are scarce, and (iii) accumulate across L11--L18, the same band where the probe and CI localize alignment. Thus the middle band must preserve this small subset; how it is approximated and protected at inference is detailed in \emph{Saliency-Guided Banded Merging} (Section~\ref{sec:tome}).

\begin{figure}[h]
\centering
\includegraphics[width=0.85\linewidth]{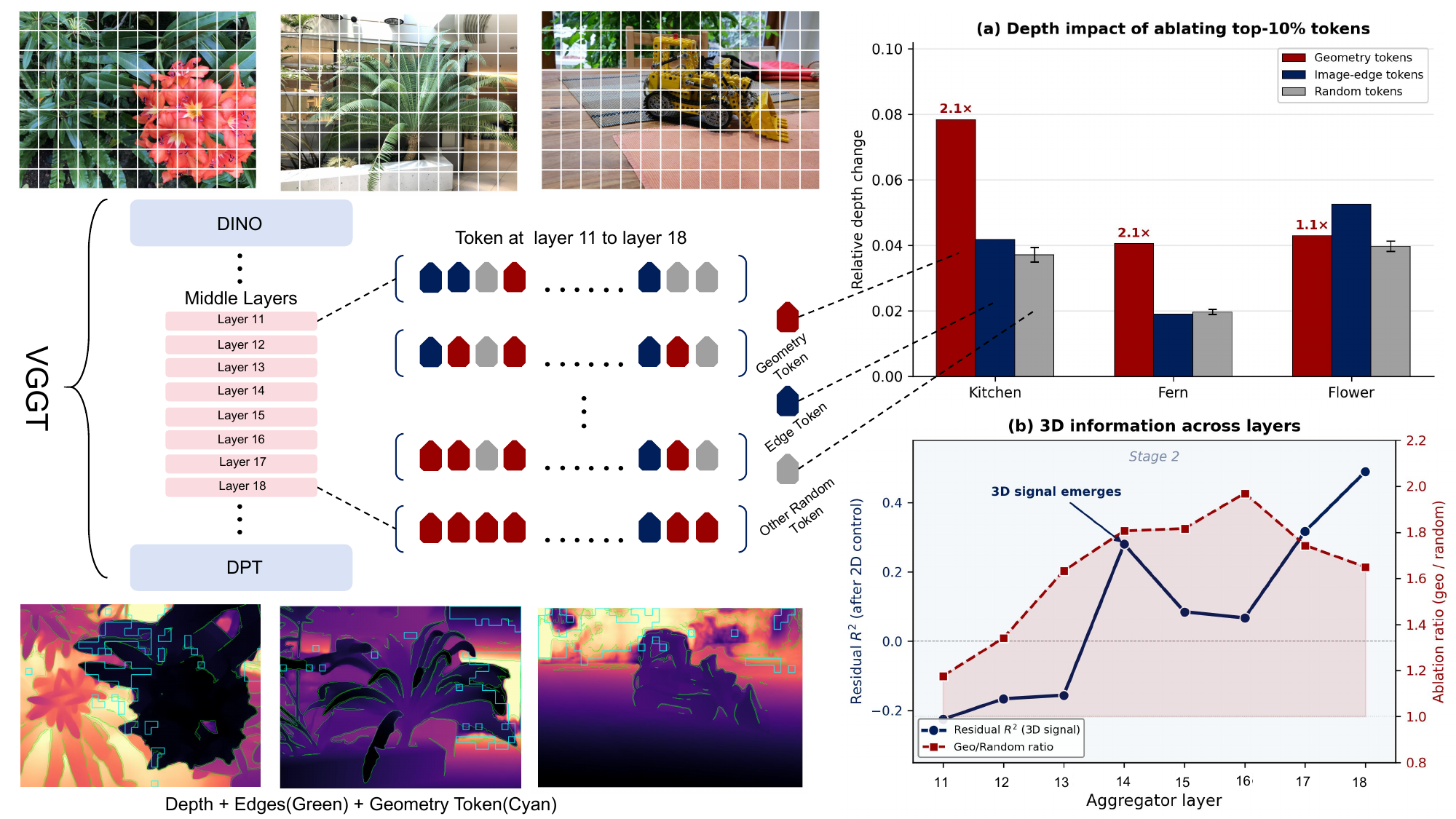}
\caption{\textbf{Heterogeneous geometry tokens and middle-band concentration.} (a)~Ablating the top-$10\%$ patch tokens by squared depth residual (after regressing out 2D cues) disrupts the depth head $2.1\times$ more than random ablation and more than a Sobel-ranked image-edge baseline on 3D-rich scenes; the gap narrows on flat scenes (Flower), confirming the score captures 3D boundaries rather than 2D texture. Bottom panels overlay geometry-token locations (cyan) on depth edges (green). (b)~Residual $R^2$ accumulates across L11--L18, the band where probe and CI localize cross-view alignment. The middle band must preserve both its rank and this small heterogeneous subset.}
\label{fig:geometry}
\vspace{-0.30cm}
\end{figure}

\paragraph{Asymmetric token roles and spectral K/V redundancy.}
Two findings on the K/V axis expose its redundancy. First, token classes play \emph{asymmetric} roles: the camera head and the pose probe both read only from the camera token, so camera and register tokens carry the final pose signal across frames---empirically, skipping deep-band global attention preserves dense reconstruction but collapses pose (Appendix~\ref{app:kv_downsample}), confirming they need a full cross-frame access path while patch tokens admit aggressive K/V compression. Second, the rank spectrum above quantifies each layer's K/V redundancy: shallow and deep layers have lower effective attention rank, even the rank-heavy middle band uses only a fraction of its capacity, and the redundancy follows a U-shape. The design and the empirical path that arrived at it are detailed in \emph{Selectively Protected K/V Downsampling} (Section~\ref{sec:kvds}) and Appendix~\ref{app:kv_downsample}, respectively.

\subsection{Saliency-Guided Banded Merging}
\label{sec:tome}
We compress the token axis: at every global block we run a bipartite merge--unmerge operator, with the merge schedule and cache schedule both governed by the three bands of Section~\ref{sec:analysis}, and the protected token subset anchored by a frozen DINOv2 \texttt{[CLS]} importance score $\Psi$. Throughout, $L = T + 5$ denotes the per-frame token count ($T$ patches plus camera and register tokens), and $L_b' = T-r_b+5$ is the post-merge count of a non-reference frame in band $b$, where $r_b$ is its merged-away patch count.

\paragraph{Per-band merge ratio.}
At each global block we apply a bipartite merge--unmerge operator adapted from ToMe and ToMeSD~\cite{tome,bolya2023token}: every non-reference frame's patch tokens are split into $\mathcal{A}_s$ and $\mathcal{B}_s$, the top-$\rho\,|\mathcal{A}_s|$ cosine-most-similar pairs are averaged, and the merged values are written back after global attention so frame attention and DPT \cite{Ranftl_2021_ICCV} taps see full resolution. The top-$\alpha$ tokens under $\Psi$ (defined below) and all reference-frame tokens are excluded from $\mathcal{A}_s$ throughout. The per-band ratios follow the rank prior:
\begin{equation}
    \rho^{(\ell)} =
    \begin{cases}
        \rho_s = 0.99 & \ell \in [1, 10] \\
        \rho_m = 0.50 & \ell \in [11, 18] \\
        \rho_d = 0.99 & \ell \in [19, 24]
    \end{cases}
    \label{eq:line_a_ratios}
\end{equation}
The U-shape, aggressive on the low-rank shallow and deep bands while conservative on the rank peak, tracks the rank analysis (Appendix~\ref{app:attention-rank}): shallow and deep layers have lower effective attention rank, while the middle band retains greater spectral complexity.

\paragraph{The protection set $\Psi$ uses frozen DINOv2 saliency.}
The middle-band emergence of 3D-boundary information (Appendix~\ref{app:geo_token}) requires per-layer protection of a small token subset; we hold out the top-$\alpha$ patches ($\alpha = 0.1$) under an importance score $\Psi$. We take $\Psi$ to be the head-averaged \texttt{[CLS]} class-attention of the frozen DINOv2 encoder's final block,
\begin{equation}
    \Psi_{p} \;=\; \frac{1}{H}\sum_{h=1}^{H}\mathrm{softmax}\!\left(\frac{q_{\mathrm{CLS}}^{(h)}(\mathbf{k}_{p}^{(h)})^{\!\top}}{\sqrt{d_{h}}}\right).
    \label{eq:dinov2_saliency}
\end{equation}
DINOv2 is trained with self-distillation and masked patch prediction~\cite{oquab2023dinov2}. Our visualization (Figure~\ref{fig:dino_spatial}) shows that its \texttt{[CLS]} attention is \emph{edge-aware but not edge-only}: it lands on salient objects rather than tracing every image edge. These object boundaries can coincide with depth discontinuities and high-frequency 2D edges, motivating high-$\Psi$ patches as an inference-time proxy for the protection target in Appendix~\ref{app:geo_token}, without ground-truth depth.

\paragraph{Three-band cache and size-aware attention.}
We add two operator-level optimizations. \emph{First}, a fixed-$\Delta$ index cache leaks across rank regimes; we instead align cache windows to band boundaries, computing matches at $\ell\in\{1,11,19\}$ and reusing them within each band, except for the rank-peak sub-range L11--L14 which is recomputed every layer. \emph{Second}, after merging each surviving key represents $g_k\ge 1$ source tokens, and the softmax must be weighted by the group-size vector $\mathbf{g}=(g_1,\dots,g_N)^{\top}$:
\begin{equation}
    \mathrm{Attn}^{\star}(\mathbf{Q},\mathbf{K},\mathbf{V};\mathbf{g}) \;=\; \mathrm{softmax}\!\left(\frac{\mathbf{Q}\mathbf{K}^{\top}}{\sqrt{d}} \,+\, \log\mathbf{g}^{\top}\right)\mathbf{V}.
    \label{eq:size_aware}
\end{equation}
Rather than implementing the bias as an attention mask~\cite{tome}, we absorb it into $\mathbf{Q},\mathbf{K}$ themselves: append a column of $1$s to $\mathbf{Q}$ and a column of $\sqrt{d}\log\mathbf{g}$ to $\mathbf{K}$, so the appended dimension contributes exactly $\log g_k$ to each logit under the standard $1/\sqrt{d}$ scale. The head dimension is zero-padded to the next FlashAttention-supported size \cite{dao2023flashattention} ($64\!\to\!80$); original output channels are recovered by slicing.

\subsection{Selectively Protected K/V Downsampling}
\label{sec:kvds}
We compress the K/V axis: at every global block, query-side updates are preserved while each non-anchor frame's K/V is restricted to a phase-shifted spatial sub-grid at a per-band rate $\sigma_b$, with cam/reg tokens, the frame-$0$ anchor, and the geometry- and edge-salient patches selected by $\Psi$ (Section~\ref{sec:tome}) held at full density. The three sub-mechanisms (rate, sub-grid, and protection) are described next.

\paragraph{Per-band K/V rate.}
The keeper density follows the same three-band rank profile as the merge ratio, but inverted: densest K/V in the rank-peak middle band, sparse on the low-rank flanks. Concretely:
\begin{equation}
    \sigma^{(\ell)} =
    \begin{cases}
        \sigma_s = 1.5 & \ell \in [1, 10] \quad \text{(kept fraction $\approx 44\%$)}\\
        \sigma_m = 1.3 & \ell \in [11, 18] \quad \text{(kept fraction $\approx 56\%$)}\\
        \sigma_d = 1.7 & \ell \in [19, 24] \quad \text{(kept fraction $\approx 36\%$)}
    \end{cases}
    \label{eq:kvds_sigma}
\end{equation}
Each band's nominal grid density is implemented on a discrete tile $(t, k)$ with $(k/t)^2 \approx 1/\sigma_b^2$. The protected-token union and reference-frame anchor add keys beyond this nominal grid density.

\paragraph{Phase-shifted per-frame sub-grid.}
A uniform sub-grid---the same in every frame---would repeatedly suppress the same image-grid coordinates. We instead \emph{phase-shift} the sub-grid by frame index: row and column offsets at frame $f$ are $(f \mathbin{\mathrm{div}} t) \bmod t$ and $f \bmod t$, where $t$ is the tile size set by $\sigma_b$. Different frames thus retain spatially offset, potentially overlapping sub-grids, diversifying the sampled image-grid positions without increasing the nominal per-frame density.

\paragraph{Selective protection: cam/reg, anchor frame, and geometry- and edge-salient patches.}
Three classes are kept at full K/V density. \textbf{Camera and register tokens} carry the asymmetric pose signal (Section~\ref{sec:analysis}) and require a full cross-frame access path. \textbf{The frame-$0$ anchor} is held at full density at every layer: it supplies a full-density reference alongside the offset grids in the other frames; in the ablation in Appendix~\ref{app:kv_downsample}, the single anchor at frame~$0$ has the highest reported AUC@30 among the tested no-anchor, two-anchor, and rotating-anchor variants. \textbf{Geometry- and edge-salient patches} share the same DINOv2 \texttt{[CLS]} score $\Psi$ as token merging (Section~\ref{sec:tome}); the top-$\alpha$ patches under $\Psi$ approximate the geometry+edge protection target (Appendix~\ref{app:geo_token}) at zero additional cost. Combined with the merging axis, the per-block cost reduces to $O(S^2 (L_b')^2 / \sigma_b^2) + O(S L^2)$, up to protected-token terms (Appendix~\ref{app:complexity}).

\section{Experiments}
\label{sec:experiments}

\subsection{Experimental Setup}

\paragraph{Datasets.}
We evaluate RegimeVGGT on dense 3D reconstruction and camera pose estimation. For dense reconstruction, we use 7 Scenes~\cite{shotton2013scene}, NRGBD~\cite{azinovic2022neural}, and ScanNet-50, a 50-scene subset of ScanNet~\cite{dai2017scannet}. For camera pose estimation, we use the Tanks \& Temples training scenes~\cite{knapitsch2017tanks}, DTU~\cite{jensen2014large}, and ScanNet-50 at varying sequence lengths.

\paragraph{Metrics.}
For 3D reconstruction, we report Accuracy (Acc$\downarrow$), 
Completeness (Comp$\downarrow$), and Normal Consistency 
(NC$\uparrow$), each as mean and median across scenes; for 
ScanNet-50 we follow FastVGGT~\cite{shen2025fastvggt} and report 
Chamfer Distance (CD$\downarrow$). For camera pose, we report 
AUC at $5^\circ/15^\circ/30^\circ$ thresholds of relative pose 
error (max of rotation and translation angular error). ScanNet-50 
long-sequence evaluation additionally uses Absolute Trajectory Error (ATE$\downarrow$), Absolute Rotation Error (ARE$\downarrow$), and Relative Pose Error in rotation and translation (RPE-rot$\downarrow$ and RPE-trans$\downarrow$).

\paragraph{Baselines and Implementation Details.}
We compare RegimeVGGT with VGGT~\cite{wang2025vggt}, VGGT$^\ast$ (the memory-efficient VGGT variant introduced by FastVGGT~\cite{shen2025fastvggt} that discards unused intermediate outputs during inference; used as our baseline), two concurrent training-free acceleration methods, FastVGGT~\cite{shen2025fastvggt} and S-VGGT~\cite{li2026s}, and two feed-forward 3D reconstruction baselines, Fast3R~\cite{yang2025fast3r} and CUT3R~\cite{wang2025continuous}. AVGGT~\cite{sun2025avggt} is compared at the design and theoretical levels in Appendix~\ref{app:avggt}; it is not part of our unified empirical benchmark. All experiments are conducted on a single NVIDIA H800 GPU with 80 GB memory at input resolution $518 \times 392$.

\subsection{3D Reconstruction}

\paragraph{Short indoor RGB-D sequences.}
Table~\ref{tab:combined_results} reports point-cloud reconstruction on 7 Scenes and NRGBD with keyframes sampled every 3 or 10 frames. This setting evaluates whether RegimeVGGT can reduce redundant global-attention computation while preserving dense local surface geometry on short indoor RGB-D sequences. On 7 Scenes, RegimeVGGT achieves the fastest inference among the VGGT-based methods under both stride settings, reducing runtime to 13.9s at stride 3 and 3.0s at stride 10, while matching the best Accuracy and maintaining competitive Completeness and Normal Consistency. On NRGBD, RegimeVGGT also gives the fastest runtime among the VGGT-based methods, with 21.8s at stride 3 and 4.2s at stride 10, while maintaining competitive reconstruction quality across all three metrics. These results indicate that RegimeVGGT reduces global-attention computation while retaining competitive dense reconstruction quality.

\begin{table}[h]
\centering
\small
\setlength{\tabcolsep}{4pt}
\caption{Quantitative results of point cloud reconstruction on the 7~Scenes and NRGBD datasets. Keyframes are sampled every 3 or 10 frames. \textit{OOM} denotes out-of-memory. Best results among VGGT-based methods are in \textbf{bold}.}
\label{tab:combined_results}
\resizebox{\linewidth}{!}{%
\begin{tabular}{l|ccccccc|ccccccc}
\toprule
& \multicolumn{7}{c|}{\textbf{Stride 3}} & \multicolumn{7}{c}{\textbf{Stride 10}} \\
\cmidrule(lr){2-8} \cmidrule(lr){9-15}
\multirow{2}{*}{\textbf{Method}}
& \multicolumn{2}{c}{Acc $\downarrow$} & \multicolumn{2}{c}{Comp $\downarrow$} & \multicolumn{2}{c}{NC $\uparrow$}
& \multirow{2}{*}{Time $\downarrow$}
& \multicolumn{2}{c}{Acc $\downarrow$} & \multicolumn{2}{c}{Comp $\downarrow$} & \multicolumn{2}{c}{NC $\uparrow$}
& \multirow{2}{*}{Time $\downarrow$} \\
\cmidrule(lr){2-3} \cmidrule(lr){4-5} \cmidrule(lr){6-7}
\cmidrule(lr){9-10} \cmidrule(lr){11-12} \cmidrule(lr){13-14}
& Mean & Med. & Mean & Med. & Mean & Med. &
& Mean & Med. & Mean & Med. & Mean & Med. & \\
\midrule
\multicolumn{15}{c}{\textbf{7 Scenes}} \\
\midrule
VGGT \textsubscript{\textit{CVPR'25}}
& \multicolumn{7}{c|}{OOM} & \multicolumn{7}{c}{OOM} \\
Fast3R \textsubscript{\textit{CVPR'25}}
& 0.045 & 0.027 & 0.047 & 0.010 & 0.605 & 0.657 & 33.0s
& 0.040 & 0.021 & 0.056 & 0.013 & 0.628 & 0.687 & 4.1s \\
CUT3R \textsubscript{\textit{CVPR'25}}
& 0.179 & 0.121 & 0.097 & 0.043 & 0.577 & 0.611 & 10.9s
& 0.041 & 0.021 & 0.029 & 0.010 & 0.640 & 0.707 & 3.2s \\
VGGT$^\ast$ \textsubscript{\textit{CVPR'25}}
& 0.018 & 0.008 & 0.028 & 0.012 & 0.600 & 0.658 & 56.0s
& 0.019 & 0.008 & 0.029 & 0.013 & 0.609 & 0.675 & 6.5s \\
FastVGGT \textsubscript{\textit{ICLR'26}}
& 0.017 & 0.008 & 0.027 & 0.012 & \textbf{0.603} & 0.662 & 18.1s
& 0.017 & 0.008 & 0.028 & 0.012 & 0.611 & 0.677 & 3.5s \\
S-VGGT \textsubscript{\textit{ICME'26}}
& 0.021 & 0.010 & \textbf{0.021} & \textbf{0.009} & 0.602 & \textbf{0.663} & 17.1s
& 0.025 & 0.012 & \textbf{0.025} & \textbf{0.011} & \textbf{0.619} & \textbf{0.688} & 4.0s \\
\midrule
\rowcolor[rgb]{0.92, 0.96, 0.99}\textbf{\methodname{}}
& \textbf{0.017} & \textbf{0.008} & 0.027 & 0.012 & 0.602 & 0.661 & \textbf{13.9s}
& \textbf{0.017} & \textbf{0.008} & 0.028 & 0.012 & 0.611 & 0.676 & \textbf{3.0s} \\
\midrule
\multicolumn{15}{c}{\textbf{NRGBD}} \\
\midrule
VGGT \textsubscript{\textit{CVPR'25}}
& \multicolumn{7}{c|}{OOM} & \multicolumn{7}{c}{OOM} \\
Fast3R \textsubscript{\textit{CVPR'25}}
& 0.074 & 0.036 & 0.024 & 0.011 & 0.588 & 0.732 & 52.0s
& 0.061 & 0.028 & 0.031 & 0.013 & 0.599 & 0.762 & 5.6s \\
CUT3R \textsubscript{\textit{CVPR'25}}
& 0.346 & 0.243 & 0.184 & 0.090 & 0.509 & 0.673 & 13.8s
& 0.132 & 0.064 & 0.056 & 0.011 & 0.599 & 0.775 & 4.3s \\
VGGT$^\ast$ \textsubscript{\textit{CVPR'25}}
& 0.028 & 0.018 & 0.018 & 0.010 & 0.657 & 0.779 & 107.3s
& \textbf{0.016} & \textbf{0.010} & \textbf{0.017} & \textbf{0.009} & 0.668 & 0.790 & 10.4s \\
FastVGGT \textsubscript{\textit{ICLR'26}}
& 0.031 & 0.020 & 0.019 & 0.010 & \textbf{0.662} & \textbf{0.788} & 29.1s
& 0.017 & 0.011 & 0.018 & 0.010 & 0.670 & 0.792 & 4.9s \\
S-VGGT \textsubscript{\textit{ICME'26}}
& 0.027 & 0.019 & \textbf{0.016} & 0.009 & 0.636 & 0.725 & 28.0s
& 0.040 & 0.029 & 0.019 & 0.011 & 0.659 & 0.761 & 4.7s \\
\midrule
\rowcolor[rgb]{0.92, 0.96, 0.99}\textbf{\methodname{}}
& \textbf{0.024} & \textbf{0.014} & 0.018 & \textbf{0.009} & 0.659 & 0.784 & \textbf{21.8s}
& 0.018 & 0.011 & 0.018 & 0.010 & \textbf{0.670} & \textbf{0.794} & \textbf{4.2s} \\
\bottomrule
\end{tabular}
}
\vspace{-3mm}
\end{table}

\paragraph{Scaling to long sequences.}
Table~\ref{tab:scannet} evaluates reconstruction on ScanNet-50 with 100--1000
input images, directly stressing the quadratic cost of global attention.
Under this ScanNet-50 protocol, the original VGGT runs out of memory at all sequence lengths, and S-VGGT also
fails at 1000 frames. At 1000 frames, \methodname{} reduces runtime to 71.5s,
compared with 477.3s for VGGT$^\ast$ and 113.5s for FastVGGT, while keeping
CD nearly unchanged. It is also the fastest VGGT-based method at 500, 300,
and 100 frames, maintaining competitive reconstruction quality across these sequence lengths. These
results show that \methodname{} preserves reconstruction quality while making
long-sequence VGGT inference substantially more scalable.

\begin{table}[h]
\centering
\small
\setlength{\tabcolsep}{4pt}
\caption{Quantitative results of point cloud reconstruction on the ScanNet-50
dataset with input sequences of 1000, 500, 300, and 100 images. \textit{OOM}
denotes out-of-memory. Best results among VGGT-based methods are in
\textbf{bold}.}
\label{tab:scannet}

\begin{tabular}{l|cc|cc|cc|cc}
\toprule
& \multicolumn{2}{c|}{\textbf{1000 frames}} & \multicolumn{2}{c|}{\textbf{500 frames}} & \multicolumn{2}{c|}{\textbf{300 frames}} & \multicolumn{2}{c}{\textbf{100 frames}} \\
\cmidrule(lr){2-3} \cmidrule(lr){4-5} \cmidrule(lr){6-7} \cmidrule(lr){8-9}
\textbf{Method} & CD $\downarrow$ & Time $\downarrow$ & CD $\downarrow$ & Time $\downarrow$ & CD $\downarrow$ & Time $\downarrow$ & CD $\downarrow$ & Time $\downarrow$ \\
\midrule
VGGT \textsubscript{\textit{CVPR'25}}
& \multicolumn{2}{c|}{OOM} & \multicolumn{2}{c|}{OOM} & \multicolumn{2}{c|}{OOM} & \multicolumn{2}{c}{OOM} \\
Fast3R \textsubscript{\textit{CVPR'25}}
& 0.684 & 299.0s & 0.701 & 73.0s & 0.711 & 26.2s & 0.723 & 3.6s \\
CUT3R \textsubscript{\textit{CVPR'25}}
& 0.786 & 26.1s & 0.774 & 14.1s & 0.775 & 8.3s & 0.767 & 2.7s \\
VGGT$^\ast$ \textsubscript{\textit{CVPR'25}}
& \textbf{0.463} & 477.3s & 0.467 & 130.2s & 0.454 & 48.7s & 0.438 & 6.8s \\
FastVGGT \textsubscript{\textit{ICLR'26}}
& 0.464 & 113.5s & 0.453 & 41.4s & 0.447 & 17.8s & \textbf{0.425} & 3.7s \\
S-VGGT \textsubscript{\textit{ICME'26}}
& \multicolumn{2}{c|}{OOM} & \textbf{0.439} & 26.1s & \textbf{0.444} & 14.3s & 0.427 & 3.2s \\
\midrule
\rowcolor[rgb]{0.92, 0.96, 0.99}\textbf{\methodname{}}
& 0.465 & \textbf{71.5s} & 0.452 & \textbf{25.6s} & 0.449 & \textbf{12.2s} & 0.430 & \textbf{3.0s} \\
\bottomrule
\end{tabular}
\vspace{-3mm}
\end{table}

\subsection{Camera Pose Estimation}
\label{sec:pose}
\begin{figure}[h]
\centering
\includegraphics[width=\linewidth]{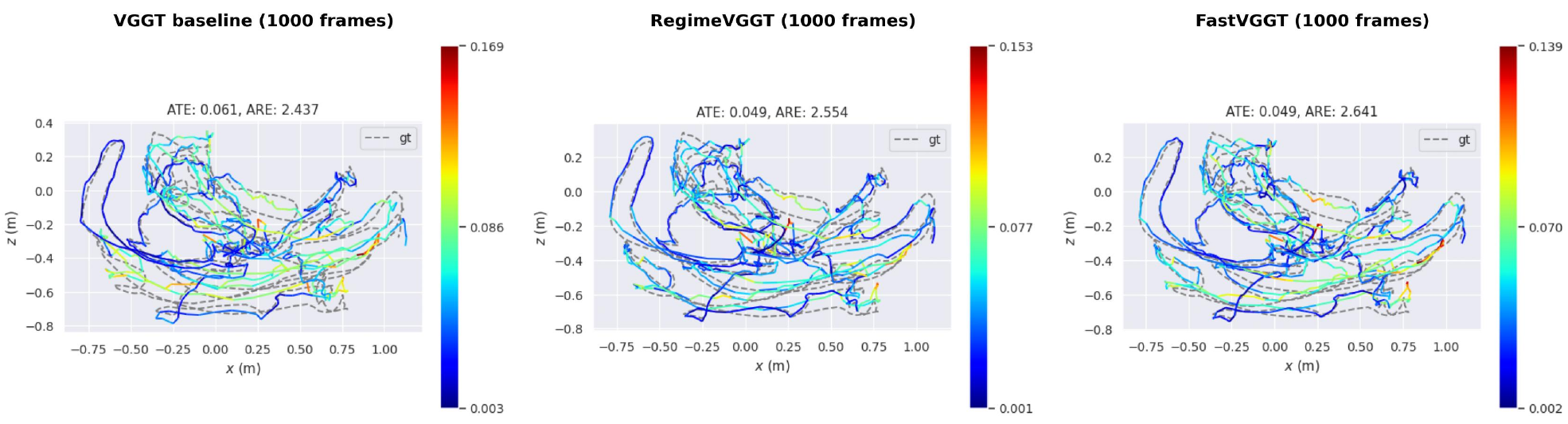}
\caption{\textbf{Predicted camera trajectories on ScanNet-50 \texttt{scene0648\_01} under 1000-frame inference.} VGGT$^\ast$ (left), \methodname{} (center), and FastVGGT (right) compared against ground truth (gray); predictions colored by per-frame ATE.}
\label{fig:pose_vis}
\vspace{-2mm}
\end{figure}

We next evaluate whether \methodname{} preserves the cross-frame geometry
needed by the camera head. Camera pose estimation directly tests whether
compression keeps the global correspondence structure that supports camera
recovery. As a qualitative example, Figure~\ref{fig:pose_vis} shows the
1000-frame trajectory on ScanNet-50 \texttt{scene0648\_01}; \methodname{}
maintains the global loop structure, consistent with the quantitative
long-sequence pose results in Table~\ref{tab:pose_scannet}.

\paragraph{Large-scale outdoor pose.}
Table~\ref{tab:pose_tnt} reports camera pose estimation on the six
Tanks~\&~Temples training scenes using per-scene AUC@30, mean AUC@30, and
total wall-clock time. \methodname{} matches FastVGGT in mean AUC@30
(90.9) while reducing total runtime from 242.2s to 111.0s. It also improves
over S-VGGT in both mean AUC@30 and runtime. \methodname{} requires
only about one fifth of the full-token VGGT$^\ast$ runtime. These results show that the
compressed global-attention computation still maintains scene-level camera
consistency on large outdoor sequences with broad viewpoint changes.

\begin{table}[h]
\centering
\small
\caption{Camera pose estimation on Tanks~\&~Temples over the six training
scenes. We report per-scene AUC@30, mean AUC@30 across scenes, and total
wall-clock inference time. Best results among VGGT-based methods are in
\textbf{bold}.}
\label{tab:pose_tnt}
\setlength{\tabcolsep}{3pt}
\begin{tabular}{l|cccccc|cc}
\toprule
\textbf{Method} & \textbf{Barn} & \textbf{Caterp.} & \textbf{Courth.} & \textbf{Ignat.} & \textbf{Meeting.} & \textbf{Truck} & \textbf{Avg.$\uparrow$} & \textbf{Time$\downarrow$} \\
\midrule
VGGT$^\ast$ \textsubscript{\textit{CVPR'25}}
& 94.4 & 92.5 & 83.8 & 92.5 & 92.8 & 93.7 & \textbf{91.6} & 552.8s \\
FastVGGT \textsubscript{\textit{ICLR'26}}
& 93.0 & 91.3 & 82.5 & 93.5 & 91.6 & 93.4 & 90.9 & 242.2s \\
S-VGGT \textsubscript{\textit{ICME'26}}
& 92.3 & 91.5 & 82.0 & 91.0 & 90.0 & 90.0 & 89.5 & 148.2s \\
\midrule
\rowcolor[rgb]{0.92, 0.96, 0.99}\textbf{\methodname{}}
& 93.1 & 91.5 & 81.2 & 93.0 & 92.6 & 93.9 & 90.9 & \textbf{111.0s} \\
\bottomrule
\end{tabular}
\vspace{-2mm}
\end{table}

\begin{table}[h]
\centering
\small
\caption{Camera pose estimation on DTU using 22 standard test scans with
48 frames per object. Best results are in \textbf{bold}.}
\label{tab:pose_dtu}
\begin{tabular}{l|ccc}
\toprule
\textbf{Method} & \textbf{AUC@30$\uparrow$} & \textbf{AUC@15$\uparrow$} & \textbf{AUC@5$\uparrow$} \\
\midrule
VGGT$^\ast$ \textsubscript{\textit{CVPR'25}}     & 94.5 & 89.0 & 67.6 \\
FastVGGT \textsubscript{\textit{ICLR'26}}  & 94.5 & \textbf{89.1} & \textbf{67.7} \\
S-VGGT \textsubscript{\textit{ICME'26}}    & 94.3 & 88.6 & 66.5 \\
\midrule
\rowcolor[rgb]{.92,.96,.99}\textbf{\methodname{}}  & \textbf{94.5} & 89.0 & 67.4 \\
\bottomrule
\end{tabular}
\vspace{-2mm}
\end{table}

\paragraph{Object-level pose.}
Table~\ref{tab:pose_dtu} reports camera pose estimation on DTU using AUC at
$30^\circ$, $15^\circ$, and $5^\circ$. On these object-level sequences, \methodname{} achieves
AUC@30 of 94.5, AUC@15 of 89.0, and AUC@5 of 67.4. These results
demonstrate that \methodname{} retains camera-relevant cross-view alignment
under training-free compression of global attention.

\paragraph{Long-sequence pose on ScanNet-50.}
Table~\ref{tab:pose_scannet} evaluates camera pose estimation on ScanNet-50
with 100, 300, 500, and 1000 input frames. This setting tests whether
\methodname{} can preserve camera consistency as the number of views grows.
At 1000 frames, \methodname{} achieves the best results across all four pose
metrics, while S-VGGT runs out of memory. At 500 frames, \methodname{} has
the lowest runtime. At
300 and 100 frames, \methodname{} also achieves the lowest VGGT-based runtime
while maintaining competitive pose accuracy. These results demonstrate
efficient camera-pose estimation across both short and long input sequences. The strongest gains appear in the
longest setting, where global-attention redundancy is most pronounced.

\begin{table}[h]
\centering
\small
\setlength{\tabcolsep}{2.4pt}
\caption{Camera pose estimation on ScanNet-50 at varying sequence lengths. We
report ATE$\downarrow$, ARE$\downarrow$, RPE-rot$\downarrow$,
RPE-trans$\downarrow$, and Time$\downarrow$. \textit{OOM} denotes
out-of-memory. Best results among VGGT-based methods are in \textbf{bold}.}
\label{tab:pose_scannet}
\begin{tabular}{l|ccccc|ccccc}
\toprule
& \multicolumn{5}{c|}{\textbf{1000 frames}} & \multicolumn{5}{c}{\textbf{500 frames}} \\
\cmidrule(lr){2-6} \cmidrule(lr){7-11}
\textbf{Method}
& ATE$\downarrow$ & ARE$\downarrow$ & R-rot$\downarrow$ & R-tr$\downarrow$ & Time$\downarrow$
& ATE$\downarrow$ & ARE$\downarrow$ & R-rot$\downarrow$ & R-tr$\downarrow$ & Time$\downarrow$ \\
\midrule
VGGT$^\ast$ \textsubscript{\textit{CVPR'25}}
& 0.117 & 5.13 & 0.96 & 0.034 & 477.3s
& 0.105 & 4.65 & 0.69 & 0.032 & 130.2s \\
FastVGGT \textsubscript{\textit{ICLR'26}}
& 0.111 & 4.17 & 0.66 & 0.029 & 113.5s
& \textbf{0.093} & \textbf{4.02} & \textbf{0.64} & \textbf{0.031} & 41.4s \\
S-VGGT \textsubscript{\textit{ICME'26}}
& \multicolumn{5}{c|}{OOM}
& 0.113 & 4.11 & 0.93 & 0.091 & 26.1s \\
\midrule
\rowcolor[rgb]{.92,.96,.99}\textbf{\methodname{}}
& \textbf{0.092} & \textbf{3.97} & \textbf{0.60} & \textbf{0.025} & \textbf{71.5s}
& 0.096 & 4.06 & 0.68 & 0.033 & \textbf{25.6s} \\
\midrule
& \multicolumn{5}{c|}{\textbf{300 frames}} & \multicolumn{5}{c}{\textbf{100 frames}} \\
\cmidrule(lr){2-6} \cmidrule(lr){7-11}
\textbf{Method}
& ATE$\downarrow$ & ARE$\downarrow$ & R-rot$\downarrow$ & R-tr$\downarrow$ & Time$\downarrow$
& ATE$\downarrow$ & ARE$\downarrow$ & R-rot$\downarrow$ & R-tr$\downarrow$ & Time$\downarrow$ \\
\midrule
VGGT$^\ast$ \textsubscript{\textit{CVPR'25}}
& 0.096 & 4.21 & 0.79 & 0.040 & 48.7s
& \textbf{0.092} & 3.95 & \textbf{1.25} & \textbf{0.056} & 6.8s \\
FastVGGT \textsubscript{\textit{ICLR'26}}
& \textbf{0.093} & 4.04 & \textbf{0.78} & \textbf{0.036} & 17.8s
& 0.093 & \textbf{3.84} & 1.26 & 0.060 & 3.7s \\
S-VGGT \textsubscript{\textit{ICME'26}}
& 0.137 & 4.29 & 1.31 & 0.122 & 14.3s
& 0.223 & 5.84 & 3.87 & 0.258 & 3.2s \\
\midrule
\rowcolor[rgb]{.92,.96,.99}\textbf{\methodname{}}
& 0.095 & \textbf{4.03} & 0.78 & 0.038 & \textbf{12.2s}
& 0.098 & 3.98 & 1.30 & 0.064 & \textbf{3.0s} \\
\bottomrule
\end{tabular}
\vspace{-2mm}
\end{table}

\subsection{Ablation Studies}
\label{sec:ablation}

Table~\ref{tab:ablation} ablates the two components of \methodname{} on Tanks~\&~Temples (Courthouse separated from the five shorter scenes). Within Saliency-Guided Banded Merging, the band-aligned cache primarily improves efficiency: removing it leaves mean AUC@30 nearly unchanged but cuts speedup from $2.78\times$ to $1.89\times$, while removing the size-bias term lowers AUC@30, especially on Courthouse. Within Selectively Protected K/V Downsampling, the reference-frame anchor and phase-shifted grid raise accuracy at unchanged runtime (removing them drops mean AUC@30 from 0.9185 to 0.9126 and 0.9132). Composing the two reaches $5.01\times$ mean and $5.79\times$ Courthouse speedup---our default \methodname{}.

\begin{table}[h]
\centering
\small
\caption{Component ablations on Tanks~\&~Temples, split by sequence length:
\emph{Long} = Courthouse alone (1106 frames), \emph{Short} = the other five
training scenes (251--410 frames each), and \emph{Mean} = all six scenes.
Speedup is wall-clock relative to VGGT$^\ast$ on the corresponding split. Higher
AUC@30 and higher speedup are better.}
\label{tab:ablation}
\setlength{\tabcolsep}{3pt}
\resizebox{0.95\linewidth}{!}{%
\begin{tabular}{ll|cc|cc|cc}
\toprule
& & \multicolumn{2}{c|}{\textbf{Long ($>$1k frames)}} 
& \multicolumn{2}{c|}{\textbf{Short}} 
& \multicolumn{2}{c}{\textbf{Mean}} \\
\cmidrule(lr){3-4} \cmidrule(lr){5-6} \cmidrule(lr){7-8}
\textbf{Config} & \textbf{Variant} 
& AUC@30$\uparrow$ & Sp.$\uparrow$ 
& AUC@30$\uparrow$ & Sp.$\uparrow$ 
& AUC@30$\uparrow$ & Sp.$\uparrow$ \\
\midrule
\multicolumn{2}{l|}{VGGT$^\ast$ (baseline)} 
& 0.8381 & 1.00$\times$ 
& 0.9319 & 1.00$\times$ 
& 0.9162 & 1.00$\times$ \\
\midrule
\multirow{3}{*}{\shortstack[l]{Saliency-Guided\\Banded Merging}}
 & \,without cache        
 & \textbf{0.8409} & 1.83$\times$ 
 & 0.9314 & 2.02$\times$ 
 & 0.9163 & 1.89$\times$ \\
 & \,without size bias   
 & 0.8225 & 2.90$\times$ 
 & 0.9311 & 2.57$\times$ 
 & 0.9064 & 2.79$\times$ \\
 & standard        
 & 0.8358 & 2.90$\times$ 
 & 0.9320 & 2.57$\times$ 
 & 0.9159 & 2.78$\times$ \\
\midrule
\multirow{3}{*}{\shortstack[l]{Selectively Protected\\K/V Downsampling}}
 & \,without anchor      
 & 0.8134 & 2.27$\times$ 
 & 0.9324 & 2.01$\times$ 
 & 0.9126 & 2.18$\times$ \\
 & \,without phase shift 
 & 0.8254 & 2.28$\times$ 
 & 0.9307 & 2.01$\times$ 
 & 0.9132 & 2.18$\times$ \\
 & standard        
 & 0.8329 & 2.28$\times$ 
 & \textbf{0.9356} & 2.01$\times$ 
 & \textbf{0.9185} & 2.18$\times$ \\
\midrule
\rowcolor[rgb]{0.92, 0.96, 0.99}RegimeVGGT & standard   
& 0.8125 & \textbf{5.79$\times$} 
& 0.9281 & \textbf{3.93$\times$} 
& 0.9088 & \textbf{5.01$\times$} \\
\bottomrule
\end{tabular}
}
\vspace{-2mm}
\end{table}

\subsection{Memory Efficiency}
\label{app:memory}

Table~\ref{tab:vram} reports peak GPU memory across the full evaluation suite. Under this evaluation protocol, VGGT exceeds 80~GB (OOM) in all reported settings, while VGGT$^\ast$ completes inference with dense global attention. Among the accelerated methods, \methodname{} has the lowest peak memory in every setting where they complete inference. Relative to VGGT$^\ast$, however, it uses 4--15\% more peak memory, so its gain is in latency rather than memory.

\begin{table*}[h]
\centering
\small
\caption{Peak GPU memory (GB) on 7~Scenes, NRGBD, and ScanNet-50, reported as the mean and max across scenes in each dataset. Best results among accelerated methods are in \textbf{bold}.}
\label{tab:vram}
\setlength{\tabcolsep}{3pt}
\resizebox{\textwidth}{!}{%
\begin{tabular}{l|cccc|cccc|cccccccc}
\toprule
& \multicolumn{4}{c|}{\textbf{Stride 3}} & \multicolumn{4}{c|}{\textbf{Stride 10}} & \multicolumn{8}{c}{\textbf{ScanNet-50}} \\
\cmidrule(lr){2-5} \cmidrule(lr){6-9} \cmidrule(lr){10-17}
& \multicolumn{2}{c}{7~Scenes} & \multicolumn{2}{c|}{NRGBD} & \multicolumn{2}{c}{7~Scenes} & \multicolumn{2}{c|}{NRGBD}
& \multicolumn{2}{c}{1000} & \multicolumn{2}{c}{500} & \multicolumn{2}{c}{300} & \multicolumn{2}{c}{100} \\
\cmidrule(lr){2-3} \cmidrule(lr){4-5} \cmidrule(lr){6-7} \cmidrule(lr){8-9}
\cmidrule(lr){10-11} \cmidrule(lr){12-13} \cmidrule(lr){14-15} \cmidrule(lr){16-17}
\textbf{Method} & Mean & Max & Mean & Max & Mean & Max & Mean & Max & Mean & Max & Mean & Max & Mean & Max & Mean & Max \\
\midrule
VGGT     & \multicolumn{4}{c|}{OOM} & \multicolumn{4}{c|}{OOM} & \multicolumn{8}{c}{OOM} \\
VGGT$^\ast$    & 24.4 & 26.3 & 30.0 & 38.6 &  8.9 &  9.5 & 10.8 & 13.2 & 42.5 & 45.3 & 23.7 & 23.7 & 15.1 & 15.1 &  6.5 &  6.5 \\
FastVGGT & 30.1 & 32.5 & 37.2 & 47.9 & 10.6 & 11.3 & 12.9 & 16.0 & 60.3 & 63.7 & 32.9 & 32.9 & 20.6 & 20.6 &  8.3 &  8.3 \\
S-VGGT   & 33.2 & 35.5 & 41.1 & 52.1 & 13.3 & 16.1 & 17.3 & 20.2 & \multicolumn{2}{c}{OOM} & 37.0 & 37.9 & 28.4 & 29.1 & 20.5 & 21.0 \\
\midrule
\rowcolor[rgb]{0.92, 0.96, 0.99}\textbf{\methodname{}} & \textbf{27.1} & \textbf{29.8} & \textbf{33.1} & \textbf{41.9} & \textbf{9.3} & \textbf{10.3} & \textbf{11.5} & \textbf{14.2} & \textbf{49.0} & \textbf{51.8} & \textbf{26.9} & \textbf{27.0} & \textbf{17.1} & \textbf{17.1} & \textbf{7.2} & \textbf{7.2} \\
\bottomrule
\end{tabular}
}
\vspace{-3mm}
\end{table*}

\section{Limitations}
\label{sec:limitations}
First, \methodname{} uses fixed per-band compression rates $(\rho_b,\sigma_b)$ chosen from scene-pooled diagnostics rather than adapting them per input. Since scene complexity varies, input-adaptive rate selection could further improve the speed--quality trade-off without changing the overall design. Second, our protected token set uses frozen DINOv2 \texttt{[CLS]} saliency, an input-fixed training-free proxy that offers competitive accuracy and memory use among the tested importance scores. If fine-tuning or additional supervision is allowed, a learned geometry-aware selector could identify geometry-critical tokens more directly and may further improve preservation under aggressive compression.

\section{Conclusion}
\label{sec:conclusion}
\methodname{} uses spectral, probing, and causal diagnostics to guide layer-wise U-shaped compression across three functional regimes in VGGT's global attention. \emph{Saliency-Guided Banded Merging} protects geometry- and edge-salient tokens; \emph{Selectively Protected K/V Downsampling} preserves spatial coverage and the pose-critical cross-frame path through phase-shifted grids, a reference-frame anchor, and uncompressed camera/register tokens. This combination compresses both token count and K/V density while keeping pretrained weights frozen, achieving a $6.7\times$ end-to-end speedup over VGGT$^\ast$ on 1000-frame ScanNet-50 at matched dense reconstruction quality.

\begin{ack}
The authors received no third-party funding or support for this work in the 36 months prior to submission. The authors also declare no financial relationships outside the submitted work during this period that could be perceived as influencing it.
\end{ack}

\bibliographystyle{plainnat}
\bibliography{refs}

\clearpage
\appendix

\section{Extended Analysis Evidence}
\label{app:extended_analysis}
This appendix collects the empirical evidence underlying \methodname{}'s two design axes---token merging and K/V downsampling---together with the experiments that motivated each design choice. Both axes rest on a layer-wise partition of VGGT's 24 aggregator layers into three functionally distinct bands (shallow, middle, deep), so we first establish that partition with three independent diagnostics---probing, causal intervention, and rank spectrum analysis~(\ref{app:layer_finding}). We then present per-axis evidence. For token merging: a geometry-token classification that defines the protection target under merging, followed by a four-way importance ablation that selects DINOv2 saliency as the deliberate scoring rule~(\ref{app:token_protection}). For K/V downsampling: the empirical path from a deep-only skip variant through asymmetric queries to the final selectively protected design (universal U-shape rate, phase-shifted per-frame grid, frame-$0$ anchor)~(\ref{app:kv_downsample}).

\subsection{Three-Band Layer Partition of VGGT's Aggregator}
\label{app:layer_finding}

\begin{figure}[h]
  \centering
  \includegraphics[width=\linewidth]{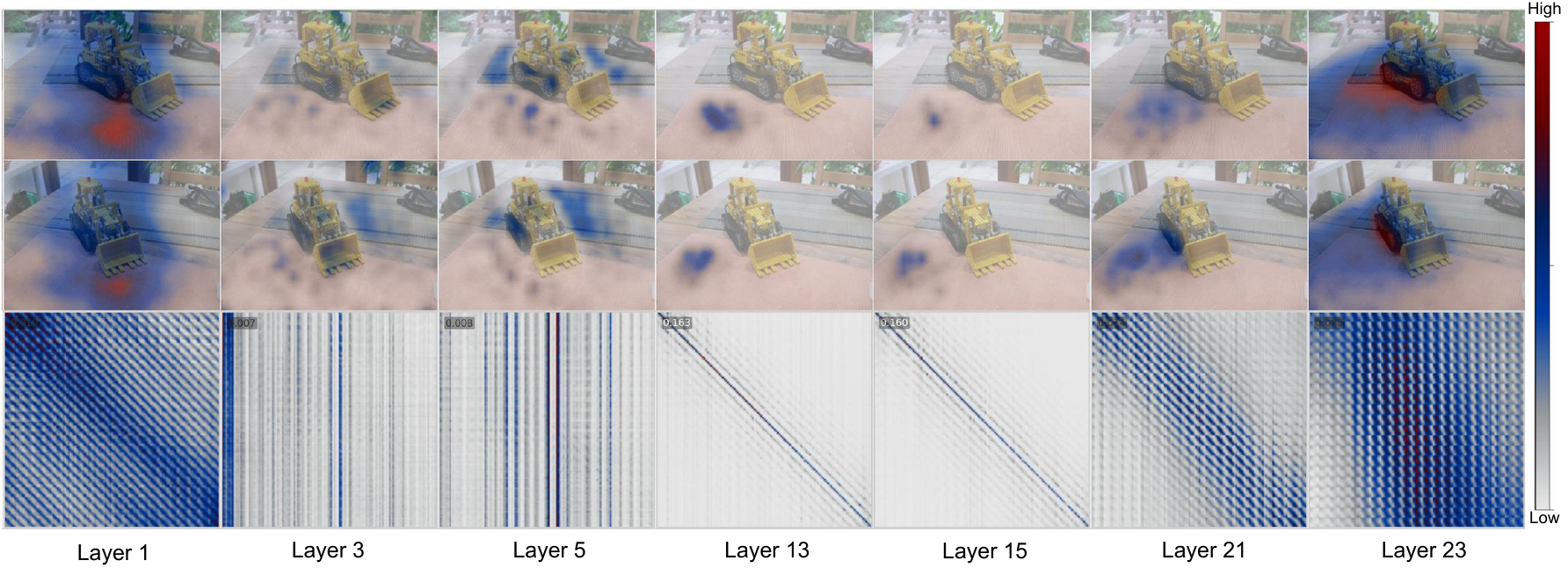}
  \caption{\textbf{Three-band layer structure in VGGT's cross-frame attention.} Top: for a fixed query patch in Frame~0, attention to all patch tokens in Frame~1 and Frame~2, visualized as spatial heatmaps. Bottom: cross-frame attention submatrix from all patch tokens in Frame~0 to all patch tokens in Frame~1.}
  \label{fig:cross_frame_attn}
\end{figure}

Section~\ref{sec:analysis} of the main paper rests on a partition of VGGT's 24 aggregator layers into three functionally distinct bands---shallow (L1--L10), middle (L11--L18), and deep (L19--L24). Figure~\ref{fig:cross_frame_attn} already exposes this partition visually: the cross-frame attention heatmaps and matrix submatrices change qualitatively across layers---diffuse and band-structured in the shallow layers (L1, 3, 5), localized with a clear near-diagonal structure in the middle layers (L13, 15), collapsed or weakly structured in the deep layers (L21, 23). To verify that this visual impression reflects an intrinsic three-band structure rather than a single layer's idiosyncrasy, we triangulate the partition with three independent diagnostics. \emph{Probing} (\ref{app:probe}) measures how much geometric information is linearly recoverable from the representations at each layer; \emph{causal intervention} (\ref{app:causal}) measures whether that information is actually utilized by the downstream prediction heads; \emph{rank spectrum analysis} (\ref{app:attention-rank}) inspects the structure of the attention map itself, free of any target.

\subsubsection{Layer-wise Probe of Geometric Information}
\label{app:probe}

\begin{figure}[h]
\centering
\includegraphics[width=0.85\linewidth]{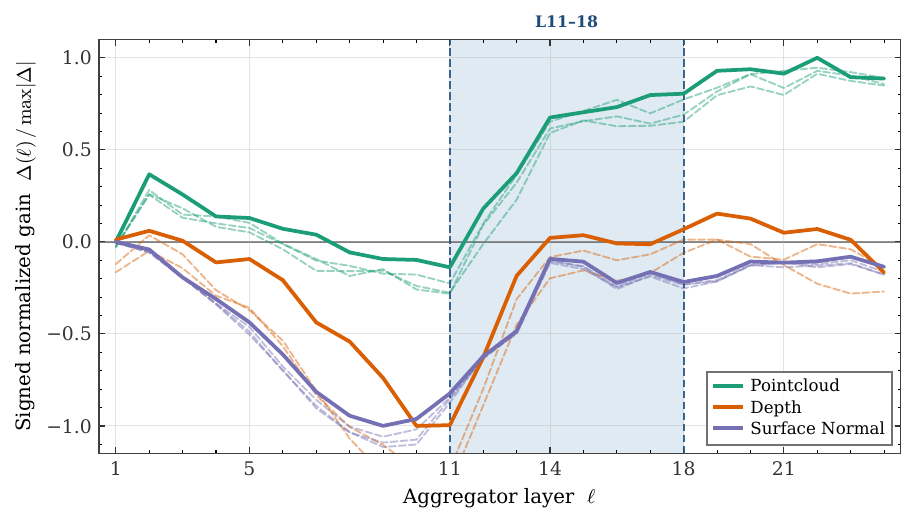}
\caption{\textbf{Geometric probe gain emerges in the middle band L11--L18.} Signed normalized gain $\Delta(\ell)/\max_j|\Delta_j|$ across VGGT's 24 aggregator layers for three dense probes (pointmap, depth, surface normal). Solid curves: default configuration with G2F removed. Dashed curves: boundary-scan variants retaining G2F through L1--$K$. Shaded region: the \emph{critical} band L11--L18.}
\label{fig:emergence}
\end{figure}

\paragraph{Closed-form ridge probe.}
At each global-attention layer $\ell$, we ask how much geometric information the layer's representations carry beyond low-level 2D image cues. We fit closed-form ridge regressors with cross-validated $\alpha$ from frozen VGGT features to four targets: depth, surface normal, camera-frame pointmap, and camera pose. The three dense targets are regressed from patch tokens, while pose is regressed from the per-frame camera token; Figure~\ref{fig:emergence} reports only the three dense patch-token probes. For the dense targets, we also fit the same ridge probe from a 6-dim control feature $\mathbf{Z}=(\text{row}, \text{col}, R, G, B, \text{Sobel})$ computed per patch, and report the control-adjusted coefficient of determination
\begin{equation}
R^2_{\text{resid}}(\ell) \;=\; R^2_{\text{full}}(\ell) - R^2_{\text{ctrl}},
\end{equation}
where $R^2_{\text{full}}$ uses the VGGT feature and $R^2_{\text{ctrl}}$ uses only the control feature. The signed per-layer gain $\Delta(\ell)$ plotted in Figure~\ref{fig:emergence} is a post-processing of this residual trajectory and is normalized by $\max_j|\Delta_j|$ for visual comparability across targets.

\paragraph{Data and robustness.}
The main probe uses 7~Scenes~\cite{shotton2013scene} and NRGBD~\cite{azinovic2022neural} with stride-10 keyframe sampling and an $80/20$ train/test split on pooled tokens. The dense targets are derived from the saved ground truth: depth is the dataset depth map, surface normal is computed from that depth map, and pointmap is obtained by unprojecting depth with the camera intrinsics. Statistical reliability is checked with a permutation test ($n{=}200$) for the dense-target $R^2_{\text{full}}$. We also record leave-one-scene-out per-scene $R^2$ values using the CV-selected $\alpha$, which we use downstream to assess cross-scene variability. We additionally run a boundary scan over global-to-frame (G2F) conversion, i.e., replacing global attention in L1--$K$ with frame attention as in AVGGT~\cite{sun2025avggt}, over several cutoffs $K$, and feed each re-extracted feature set through the same probe.

\paragraph{Results.}
Figure~\ref{fig:emergence} reports the signed, max-normalized per-layer gain
$\Delta^{(\ell)}/\max_j|\Delta^{(j)}|$ for the three dense per-patch probes:
pointmap, depth, and surface normal. The solid curves denote the default
configuration used in our final model, which applies no G2F conversion. The dashed
curves denote boundary-scan variants that retain G2F through layers
L1--$K$ for several tested cutoffs $K$. Across the tested cutoffs, retaining
shallow G2F yields weaker aggregate dense-geometry probe signals than the
default no-G2F configuration, indicating that early G2F suppresses rather
than preserves useful dense-geometry accumulation. We therefore do not use G2F
conversion in the final model. This transition is consistent with
Figure~\ref{fig:cross_frame_attn}, where attention changes from diffuse,
weakly structured shallow patterns to localized near-diagonal cross-frame structure in the middle layers.

\subsubsection{Causal and Structural Diagnostics of Cross-View Alignment}
\label{app:causal}

\paragraph{From availability to utilization.}
The probe in \ref{app:probe} establishes that geometric information is linearly \emph{available} in the representations of L11--L18, but does not confirm that the prediction heads actually \emph{use} it. We close this gap with two complementary tests on every global-attention layer: a \emph{causal} test that asks whether perturbing the layer's patch-key correspondence changes the downstream output, and a \emph{structural} test that asks whether the attention pattern itself looks like geometric correspondence matching. Either test in isolation is only suggestive---a layer's attention can look sharp without driving the output, or perturbations may matter for unrelated reasons---but their \emph{convergence} pins a layer as a genuine cross-view alignment site.

\paragraph{Experiment.}
We compute four per-layer readouts on the same scenes and model configuration, using 7~Scenes~\cite{shotton2013scene} capped at 16 frames per sequence to bound the $O(S^2)$ aggregator memory. \emph{(i)} For the causal influence $\mathrm{CI}(\ell)$, we install a forward hook on each layer's $\mathbf{QKV}$ projection that overwrites the keys $\mathbf{K}$ of all patch tokens with a single random permutation of patch positions across the entire frame stack, leaving $\mathbf{Q}$ and $\mathbf{V}$ unchanged; we then forward the model and record
\begin{equation}
\mathrm{CI}(\ell) \;=\; \frac{\|\mathrm{head\_out}_{\mathrm{perm}} - \mathrm{head\_out}_{\mathrm{clean}}\|_2}{\|\mathrm{head\_out}_{\mathrm{clean}}\|_2}
\end{equation}
for the depth and camera heads, averaged over $n_{\text{perm}}{=}5$ permutation seeds. \emph{(ii)} For the structural diagnostics, we approximate attention scores from qkv-projected $(\mathbf{Q},\mathbf{K})$ on $n_{\text{query}}{=}64$ sampled patch queries per frame (flash-attention does not expose softmax weights) and report the normalized entropy $H = -\sum_k a_k \log a_k / \log NL \in [0, 1]$, plotted as $1{-}H$ so that high values mean sharp attention, and the \emph{diagonal ratio}: the fraction of cross-frame attention mass landing within $\pm 2$ patches of each query's spatial position.

\begin{figure}[h]
\centering
\includegraphics[width=\linewidth]{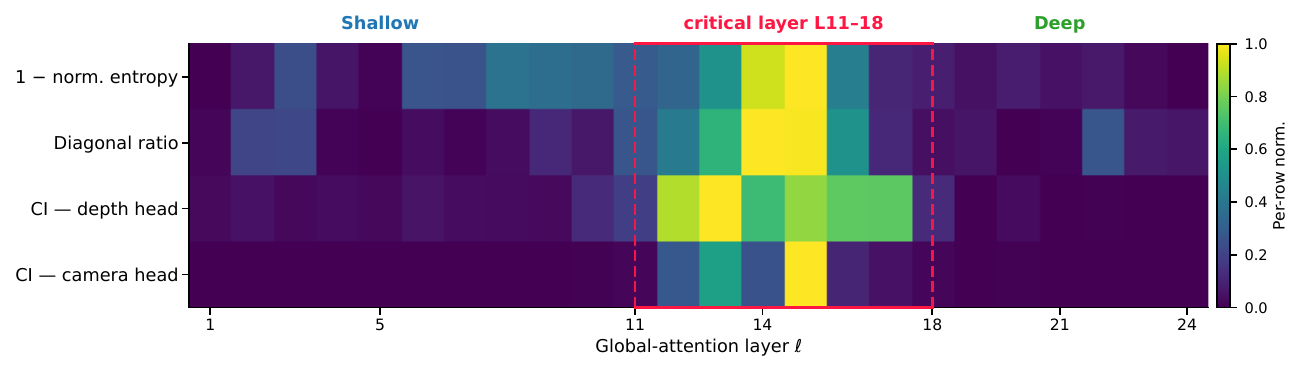}
\caption{\textbf{Four diagnostics co-localize cross-view alignment to L11--L18.} Rows: $1{-}\mathrm{norm.~entropy}$ (sharpness), diagonal ratio (cross-frame correspondence), causal influence on depth head, causal influence on camera head; each row per-row normalized. The critical band L11--L18 (red box) is bright on all four rows; the camera CI peaks near L15.}
\label{fig:ci_per_layer}
\end{figure}

\paragraph{Analysis.}
Figure~\ref{fig:ci_per_layer} reports the four metrics, each per-row normalized, across the 24 global-attention layers. The four rows collapse onto the same partition. The shallow band (L1--L10) is uniformly dim: attention is unfocused, unaligned with the cross-frame correspondence diagonal, and key permutation produces no measurable change in either head. The deep band (L19--L24) is also dim across the causal rows and shows little cross-frame diagonal structure, suggesting that these layers are not the main geometric correspondence-building stage. The critical band L11--L18 (red box) is bright across all four rows: attention is sharply concentrated and lies along the cross-frame correspondence diagonal, while key permutation shifts the outputs measurably, with the camera CI peaking at around L15 and the depth CI at around L14. The two structural and two causal signals \emph{co-localize} to L11--L18, supplying the convergence requirement set up above: this is where cross-view geometric attention actually does its work. We interpret the gap between camera and depth CI as partly reflecting head architecture: the camera head concentrates a low-dimensional pose prediction into a single bottleneck token, while depth is per-patch with redundancy that buffers single-layer disruptions.

\subsubsection{Rank Spectrum of Global Attention}
\label{app:attention-rank}

\paragraph{Target-free structural diagnostic.}
The probe and causal-intervention results above localize the same middle band from two end-to-end perspectives, but both rely on ground-truth targets or prediction-head outputs. We therefore ask whether the same layer-wise partition is already visible in the global-attention computation itself, and analyze the rank spectrum of each layer's attention probability matrix.

\paragraph{Sampled attention-rank: metric and setup.}
To probe the layer-wise partition without relying on labels or prediction heads, we examine the rank spectrum of each layer's global-attention probability matrix. The rank reflects the dimensionality of the cross-token interactions a layer actually exercises: a low-rank matrix indicates that all queries attend through a small shared subspace---routing, copying, or smoothing---whereas a high-rank matrix indicates diverse, query-specific aggregation, the regime in which non-trivial cross-view computation can occur. Since the full token-by-token attention matrix is too large to decompose directly on long sequences, for each layer $\ell$ and head $h$ we sample $m$ query rows and compute their attention weights over all $N=SP$ tokens, producing an $m \times N$ submatrix. We then compute its singular values and measure the smallest rank required to explain 95\% of the spectral energy:
\[
r_{95}^{(\ell,h)}
=
\min
\left\{
r:\;
\frac{\sum_{i=1}^{r}\sigma_i^2}
{\sum_i \sigma_i^2}
\ge 0.95
\right\}.
\]
We report the normalized rank ratio $r_{95}^{(\ell,h)}/m$, averaged over heads and scenes. We evaluate the metric on scenes from DTU, ScanNet, and Tanks~\&~Temples, covering object-centric, indoor, and large-scale outdoor settings. Unless otherwise stated, we use $m=256$ sampled query rows and at most 32 frames per scene, and aggregate $r_{95}^{(\ell,h)}/m$ over all heads at each layer.

\paragraph{Consistent layer boundaries.}
This target-free spectral diagnostic provides an independent structural confirmation of the three-band partition. The middle band is not only where geometric information is available and causally used; it is also where the global-attention probability map has the highest spectral complexity. This indicates that L11--L18 contains the richest cross-view interaction structure, consistent with its role as the main alignment regime. By contrast, the shallow and deep bands have lower spectral complexity, although for different reasons: the shallow band has not yet developed strong cross-view matching structure, while the deep band has largely collapsed into a redundant refinement regime. Thus, the same L11 and L18 boundaries emerge from representation probing, causal use, and the intrinsic spectrum of global attention itself. Because this spectrum is measured on the attention probability map, the low-rank flanks indicate redundancy in how queries read from the global K/V set, while the rank-heavy middle band requires denser K/V support to preserve its richer cross-view correspondence structure.

\begin{figure}[h]
\centering
\includegraphics[width=0.9\linewidth]{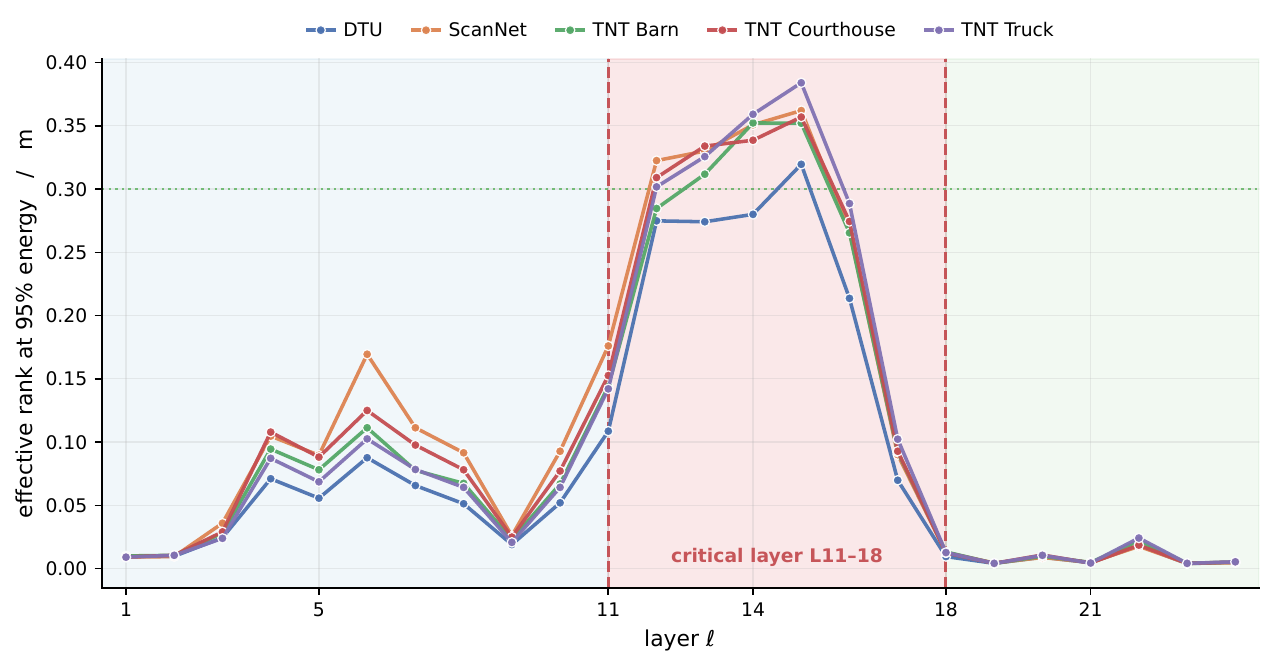}
\caption{\textbf{Effective attention rank peaks in the middle band across the five evaluated scenes.} Effective rank of the global-attention matrix vs.\ layer depth, on five scenes (TNT Courthouse, TNT Barn, TNT Truck, DTU, ScanNet); $m{=}256$ sampled query rows per layer. The middle band $\ell{\in}[11,18]$ has higher effective rank than the shallow and deep bands. The green dotted line marks a normalized rank ratio of $0.30$.}
\label{fig:rank_spectrum}
\end{figure}

\subsection{Token-Level Heterogeneity and Protection Targets}
\label{app:token_protection}

The layer-wise analyses in Appendix~\ref{app:probe}--\ref{app:attention-rank} show where VGGT can be compressed across depth, but token merging also requires a second question: \emph{which patch tokens should survive the merge?} This is not a uniform token-reduction problem. In VGGT, each patch token remains spatially tied to dense 3D prediction through the DPT heads, so merging the wrong tokens can erase local structure that cannot be recovered later.

Figure~\ref{fig:geometry} in the main paper motivates this token-level view. It shows that the middle band L11--L18 is not only the regime where cross-view geometric information emerges, but also the regime where a small heterogeneous subset of patch tokens carries disproportionate 3D information. Ablating the top-$10\%$ geometry-boundary tokens disrupts the depth head much more than ablating random tokens, while the residual 3D signal accumulates across the same L11--L18 window identified by the probe and causal-intervention analyses. This appendix expands that observation in two steps. Appendix~\ref{app:geo_token} defines the token classes that should be protected during token merging. Appendix~\ref{app:importance} then compares practical importance scores for selecting such protected tokens during inference.

\subsubsection{Geometry and Edge Tokens as Protection Targets for Token Merging}
\label{app:geo_token}

\paragraph{Motivation.}
Token merging removes redundancy by combining similar patch tokens, but dense 3D reconstruction is sensitive to local spatial failures. Smooth image regions often contain many redundant tokens and can tolerate merging, whereas tokens near depth changes, object boundaries, or high-frequency spatial structures are more fragile. We therefore separate the protection target into two complementary token types: \emph{geometry tokens}, which denote patch locations whose depth cannot be explained by local 2D cues alone and therefore serve as a proxy for 3D-critical regions, and \emph{edge tokens}, which mark high-frequency image boundaries that are vulnerable to merging.

\paragraph{Geometry tokens.}
We define geometry tokens through a residual-depth criterion, similar to Appendix~\ref{app:layer_finding}. For each patch token, we first regress the ground-truth patch depth from low-level 2D cues: normalized image coordinates, patch RGB values, and Sobel edge magnitude. This control regression explains the portion of depth that can already be predicted from image position, color, and local edge strength. We then rank tokens by the squared residual of this regression,
\begin{equation}
e_i =\left(y^{\mathrm{depth}}_i - \hat{y}^{\mathrm{2D}}_i \right)^2,
\end{equation}
and define the top-$10\%$ residual tokens as geometry tokens. A high residual means that the token's depth cannot be explained by local 2D cues alone, so it is more likely to correspond to genuine 3D structure such as depth discontinuities or scene-layout changes.

\paragraph{Edge tokens.}
We define edge tokens independently using Sobel magnitude on the input image. These tokens capture high-frequency spatial structures such as object contours, texture boundaries, and possible depth edges. Edge tokens are not equivalent to geometry tokens: some strong image edges are appearance-only texture, while some geometry tokens correspond to 3D structure that is not the strongest 2D edge. The two sets therefore provide complementary protection targets. Geometry tokens target patch locations associated with residual 3D information, while edge tokens target spatial boundaries that are easy to blur under merging. Together, they define the desired protection target for token merging, whereas the practical policy must approximate this target using inference-time signals that do not rely on ground-truth depth.

\paragraph{Functional validation.}
To verify that geometry tokens are not an artifact of the residual definition, we perform a full layer-wise ablation experiment by applying the same token-ablation procedure across the aggregator layers. For each frame, we ablate the top-$10\%$ geometry tokens and compare against ablating the same fraction of randomly selected patch tokens. We measure the relative change in VGGT's DPT depth output outside the ablated regions, so that the metric reflects how perturbing a selected token set propagates to the surrounding dense geometry rather than only the directly masked locations. As reported in Figure~\ref{fig:geometry}, geometry-token ablation causes about $2.1\times$ larger depth disruption than random ablation on 3D-rich scenes. On geometrically flatter scenes, such as Flower, the gap narrows to about $1.1\times$, consistent with the smaller number of strong depth discontinuities in such scenes. This confirms that the residual-depth criterion identifies a functional token class whose removal disproportionately affects dense geometry.

\paragraph{Implication for token merging.}
These observations define the target that token merging should preserve. A safe merge policy should preferentially keep patch tokens associated with residual 3D information and tokens lying on sharp spatial boundaries, while allowing redundant tokens in smooth regions to be merged. The next section evaluates practical inference-time importance scores for approximating this protection target.

\subsubsection{Importance Ablation}
\label{app:importance}

Given the protection target defined above, we next ask which inference-time per-token score should be used to approximate this target in token merging.

We compare four candidate per-token importance scores $\Psi$ for the protection set used by token merging and the hybrid compression scheme: (i) DINOv2 [CLS] class-attention from the last DINOv2 block, head-averaged; (ii) aggregator frame-attention saliency at L17; (iii) Sobel edge magnitude on the input image; (iv) $\ell_2$ norm of the L17 patch token. For each scoring rule, we run the full Saliency-Guided Banded Merging pipeline under the canonical configuration and report mean Acc / NC / Time / VRAM on 7 Scenes and NRGBD at strides 3 and 10 (Table~\ref{tab:ablation_importance}).

\begin{table*}[h]
\centering
\small
\caption{\textbf{DINOv2 \texttt{[CLS]} attention offers competitive accuracy and memory use.} Importance scoring variants on 7~Scenes (18 scenes) and NRGBD (8 scenes) at stride 3/10. Memory reported as Mean\,/\,Max across scenes (GB).}
\label{tab:ablation_importance}
\setlength{\tabcolsep}{3pt}
\begin{tabular}{l|cccc|cccc}
\toprule
\multicolumn{9}{c}{\textbf{Stride 3}} \\
\midrule
& \multicolumn{4}{c|}{7~Scenes} & \multicolumn{4}{c}{NRGBD} \\
\cmidrule(lr){2-5} \cmidrule(lr){6-9}
\textbf{Importance} & Acc$\downarrow$ & NC$\uparrow$ & Time$\downarrow$ & VRAM$\downarrow$
                    & Acc$\downarrow$ & NC$\uparrow$ & Time$\downarrow$ & VRAM$\downarrow$ \\
\midrule
\texttt{frame\_attn}
 & 0.017 & \textbf{0.603} & 14.19 & 32.7\,/\,72.4
 & 0.019 & 0.663 & 22.73 & 38.2\,/\,49.1 \\
\texttt{sobel}
 & 0.017 & 0.601 & 13.99 & 30.6\,/\,33.3
 & 0.019 & 0.661 & 22.55 & 38.2\,/\,49.1 \\
\texttt{token\_norm}
 & 0.017 & 0.600 & \textbf{13.94} & 30.6\,/\,33.3
 & 0.031 & 0.659 & 22.52 & 38.2\,/\,49.1 \\
 \rowcolor[rgb]{.92,.96,.99}
 \texttt{dino\_attn} (ours)
 & \textbf{0.017} & 0.602 & 13.96  & \textbf{30.6\,/\,33.3}
 & \textbf{0.019} & \textbf{0.663} & \textbf{22.52} & \textbf{38.2}\,/\,\textbf{49.1} \\
\midrule
\multicolumn{9}{c}{\textbf{Stride 10}} \\
\midrule
& \multicolumn{4}{c|}{7~Scenes} & \multicolumn{4}{c}{NRGBD} \\
\cmidrule(lr){2-5} \cmidrule(lr){6-9}
\textbf{Importance} & Acc$\downarrow$ & NC$\uparrow$ & Time$\downarrow$ & VRAM$\downarrow$
                    & Acc$\downarrow$ & NC$\uparrow$ & Time$\downarrow$ & VRAM$\downarrow$ \\
\midrule
\texttt{frame\_attn}
 & 0.017 & 0.611 & 2.97  & 10.9\,/\,11.6
 & \textbf{0.018} & 0.670 & 4.15  & 13.3\,/\,16.4 \\
\texttt{sobel}
 & 0.017 & 0.609 & 2.94  & 10.8\,/\,11.6
 & 0.021 & 0.666 & 4.10  & 13.1\,/\,16.4 \\
\texttt{token\_norm}
 & 0.017 & 0.611 & 2.92  & 10.8\,/\,11.6
 & 0.028 & 0.665 & 4.10  & 13.1\,/\,16.4 \\
\rowcolor[rgb]{.92,.96,.99}
\texttt{dino\_attn} (ours)
 & \textbf{0.017} & \textbf{0.611} & \textbf{2.92} & \textbf{10.8\,/\,11.6}
 & 0.019          & \textbf{0.673} & \textbf{4.10} & \textbf{13.1\,/\,16.4} \\
\bottomrule
\end{tabular}
\vspace{-2mm}
\end{table*}

\begin{figure}[t]                                                      \centering                                             
    \includegraphics[width=\linewidth]{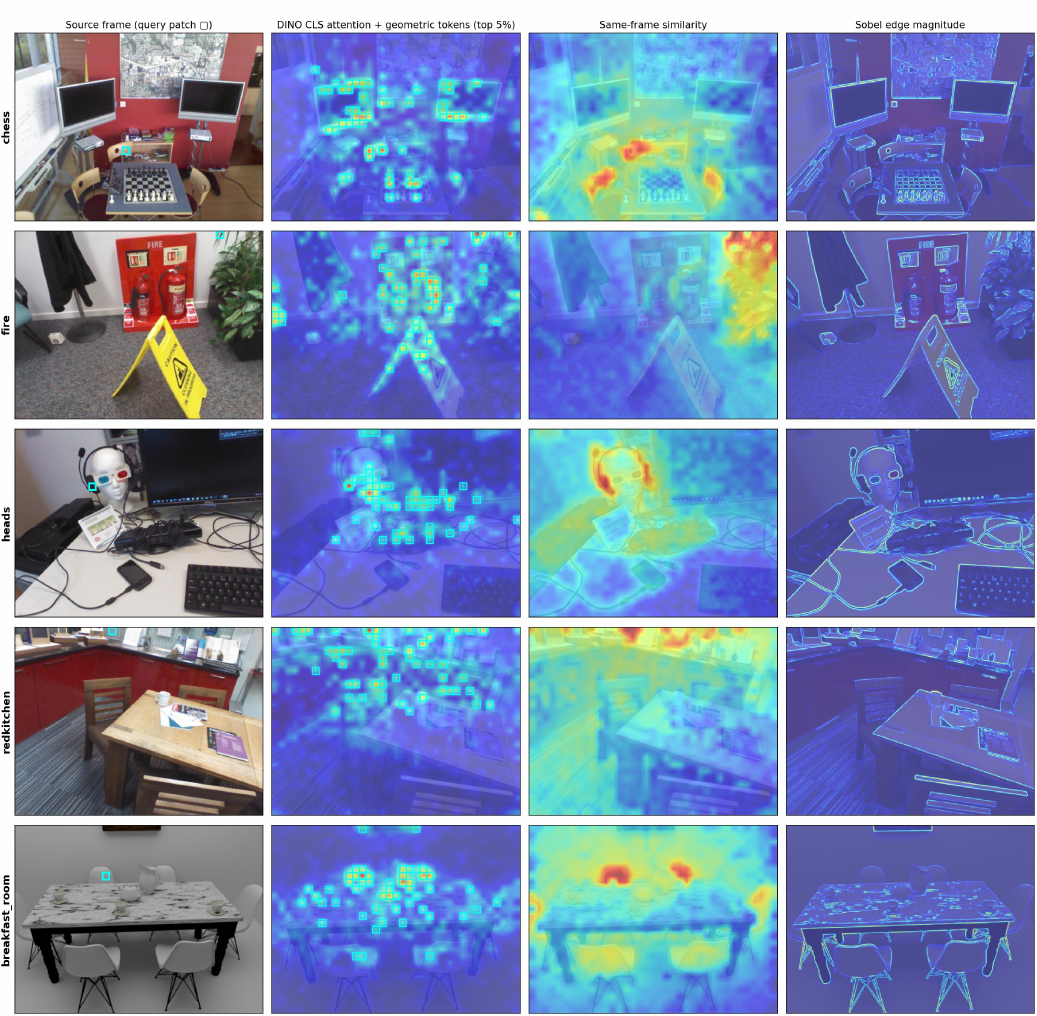}  \caption{\textbf{DINOv2 \texttt{[CLS]} attention is edge-aware but not edge-only, motivating its use as the protection signal in \methodname{}.} Across five scenes (rows): \textbf{(col~1)} source frame with the top-1 CLS-attention patch outlined in cyan;
  \textbf{(col~2)} DINOv2 \texttt{[CLS]} attention map from the last DINOv2 block, with the top-5\%
   patches outlined in cyan for visualization; the inference policy protects the top-10\% under this saliency score (Section~\ref{sec:tome}), which serves as a proxy for the residual-defined geometry-token target; \textbf{(col~3)} cosine
  similarity between the query patch and every other patch in the same frame;
  \textbf{(col~4)} Sobel edge magnitude.}
    \label{fig:dino_spatial}
  \end{figure}

The DINOv2 score is computed once per input image and is therefore \emph{input-fixed}: it does not depend on the aggregator's intermediate state, does not interact with the merge schedule, and does not accumulate noise across layers. This distinguishes it from frame-attention and token-norm scores, which drift as earlier merging changes the token representations they read from. Figure~\ref{fig:dino_spatial} further shows DINOv2 attention is \emph{not edge-only}: it lands on salient objects rather than tracing every image edge, so unlike Sobel, which is input-fixed but captures only local 2D edge strength, it approximates the geometry- and edge-sensitive protection target defined above. We therefore use \texttt{dino\_attn} as the default scoring rule for $\Psi$.

\subsection{Empirical Path to Selectively Protected K/V Downsampling}
\label{app:kv_downsample}

This appendix documents the empirical path that produced our final K/V downsampling design and the failure modes that motivated each step along the way. The final configuration (Appendix~\ref{app:asym_partial}) keeps the camera and register tokens at full K/V density to preserve the cross-frame pose path, and partially updates the remaining patch tokens through phase-shifted K/V downsampling applied universally across the global stack, with a frame-$0$ anchor preserved at full density and the per-band rate set by the rank budget reported in \ref{app:attention-rank}. The configuration was reached only after two more aggressive deep-band-only variants---direct skip of the deep global attention (Appendix~\ref{app:asym_skip}) and asymmetric queries restricted to the deep band (Appendix~\ref{app:asym_strict})---each broke pose in a distinct way; we detail the variants and their failure modes below, with quantitative comparison in Table~\ref{tab:asym_path}.

\subsubsection{Step 1 --- Skipping the deep global attention}
\label{app:asym_skip}

Our first instinct was the simplest one consistent with the analysis: if deep global attention has low effective rank and contributes little to dense reconstruction, just skip it. We replaced the global block at L19--L24 with a no-op, so each frame in the deep band is processed in isolation through the existing frame-attention path, and left every other layer (including all of L1--L18) unchanged. The reconstruction numbers held---confirming that deep global attention is not where dense outputs live---but the camera-pose pipeline collapsed on long sequences (Table~\ref{tab:asym_path}). Without deep-band global updates, the camera token at the top of the aggregator stops receiving fresh cross-frame information in those final six layers, and the resulting pose drift accumulates with sequence length until the trajectory becomes unusable. Deep global attention, despite contributing little to reconstruction, carries the cross-frame signal that lets the camera token resolve drift; we cannot remove that path wholesale.

\subsubsection{Step 2 --- Asymmetric queries inside the deep band}
\label{app:asym_strict}

Step~1 told us we need \emph{some} cross-frame computation in the deep band but not all of it---the question was which subset of tokens still needs cross-frame access. By architecture, VGGT's camera head reads from the camera token, and the pose probe in \ref{app:probe} is regressed directly from that token (with the register tokens traveling alongside it across frames); cam and register are by design the tokens carrying the final pose signal through the aggregator. We therefore tried a strict asymmetric variant restricted to L19--L24: only the camera and register tokens issue cross-frame queries in the deep band, while patches pass through the deep global block untouched (they still update through within-frame attention, and they continue to use the full global attention in L1--L18 as before). At short sequence lengths this looked like the answer---pose drift was effectively absent and ATE matched the baseline. The failure surfaced only as we scaled to longer sequences, where drift returned and grew with the number of frames. The mechanism was more subtle than Step~1: at L19--L24 the camera and register tokens still query patch K/V across frames, but patches frozen inside the deep global block stop tracking the cross-frame consensus the rest of the aggregator is building, so cam/reg tokens read from increasingly stale patch evidence as the sequence grows. The tested deep-only variants (Steps~1 and~2) therefore did not provide a satisfactory speed--pose trade-off on long sequences, motivating a design that preserves patch updates while compressing K/V across all bands.

\subsubsection{Step 3 --- Selectively Protected K/V Downsampling}
\label{app:asym_partial}

Step~2's failure pinpointed the smallest fix that could rescue it: patches need to update through global attention, but their full K/V is not necessary---replacing it with a downsampled K/V leaves all attention paths intact (cam/reg still query, patches still query, patches still get updated) while reducing the cross-frame compute. The final design factors into three complementary mechanisms described next.

\paragraph{Downsampling.} The mechanism, once written down, has no built-in dependence on the layer it lives in: K/V downsampling at deep is the same operation as K/V downsampling at middle or shallow, only with a different rate. Restricting it to L19--L24 would leave free compression on the table at every other layer, where the rank measurements in \ref{app:attention-rank} say there is also slack to take. We therefore apply the same K/V downsampling method at every layer, with the per-band rate set by the rank profile; the specific $\sigma_b$ values and per-band kept fractions are given in Section~\ref{sec:kvds}. At every layer all tokens participate as queries---including patches, so the staleness mechanism behind Step~2 disappears. The K/V side is selectively protected: camera and register tokens, together with the geometry and edge tokens identified by $\Psi$, are kept at full density.

\paragraph{Phase shift.} Within a single frame, downsampling at rate $\sigma$ keeps patch K/V on a regular sub-grid whose kept fraction is approximately $1/\sigma^2$, instantiated on a discrete tile $(t, k)$ with $k/t \approx 1/\sigma$ so that $(k/t)^2$ matches the continuous target. If every frame retained the same sub-grid, the cross-frame K/V union would inherit the same spatial blind spots, and queries arriving at any frame would systematically miss the suppressed coordinates. We avoid this by \emph{phase-shifting} the grid as a function of the frame index: at frame $f$ with tile size $t$ (set by $\sigma$), the row and column offsets are $(f \mathbin{\mathrm{div}} t) \bmod t$ and $f \bmod t$, so different frames retain different sub-grids. The offset grids diversify the image-grid positions represented across frames without increasing nominal per-frame sampling density. This coverage argument concerns sampled image coordinates; it does not guarantee that every 3D correspondence is retained.

\paragraph{Anchor frame.}
On top of phase-shifted per-frame downsampling, we keep a single anchor frame at full K/V density, fixed at frame~$0$. This supplies a fully resolved reference alongside the sampled context from the other frames. Table~\ref{tab:anchor_ablation} compares four anchor configurations on Tanks~\&~Temples 6-scene pose at a uniform $\sigma_b{=}1.5$. The single anchor at frame~$0$ has the highest reported AUC@30 and speedup among the tested variants. Adding a second anchor does not improve the reported AUC, while rotating the anchor lowers it. These observations support using a fixed reference anchor in this setting, without establishing that additional anchors can never help.

\begin{table}[h]
\centering
\small
\caption{\textbf{Single anchor at frame~0 outperforms other anchor configurations.} Anchor variants on Tanks~\&~Temples 6-scene pose, with a uniform $\sigma_b{=}1.5$ across bands and variants to isolate the anchor configuration (the U-shape rates of the final design are not used here).}
\label{tab:anchor_ablation}
\setlength{\tabcolsep}{4pt}
\begin{tabular}{l|cc}
\toprule
\textbf{Variant} & \textbf{AUC@30$\uparrow$} & \textbf{Speedup} \\
\midrule
no anchor (phase-shift only)              & 0.913 & $2.05\times$ \\
two anchors (frames 0, $S/2$)             & 0.918 & $1.99\times$ \\
rotating anchor (frame $\ell \bmod S$)    & 0.911 & $2.06\times$ \\
\rowcolor[rgb]{.92,.96,.99}\textbf{single anchor at frame~0 (ours)}  & \textbf{0.919} & \textbf{$2.09\times$} \\
\bottomrule
\end{tabular}
\end{table}

Putting these pieces together, the final K/V downsampling configuration applies a phase-shifted per-frame sub-grid at every layer, with cam/reg tokens, the frame-$0$ anchor, and the $\Psi$-selected patches kept at full density. Combined with \emph{Saliency-Guided Banded Merging} (Section~\ref{sec:tome}), this configuration produces the $6.7\times$ speedup over VGGT$^\ast$ on 1000-frame ScanNet-50 reported in the main paper. Table~\ref{tab:asym_path} quantifies the cost of each shortcut on Tanks~\&~Temples pose: skipping deep global wholesale (Step~1) collapses AUC@30 from $0.9162$ to $0.6330$, the deep-band asymmetric variant (Step~2) only recovers to $0.7965$, while selective K/V protection (Step~3) matches the baseline at $0.9185$ with a $2.18\times$ speedup.

\begin{table}[h]
\centering
\small
\caption{\textbf{Step~3 outperforms the deep-band-only variants on Tanks~\&~Temples pose.} Row~1 is the VGGT$^\ast$ baseline; rows~2--4 are the three variants discussed in \ref{app:asym_skip}--\ref{app:asym_partial}. Mean AUC@30 across the 6 training scenes (kf\_every$=$1, same protocol as Table~\ref{tab:pose_tnt}); total wall-clock; speedup over VGGT$^\ast$. Step~1 collapses pose ($-0.28$\,AUC@30); Step~2 only partially recovers ($-0.12$); Step~3 matches the baseline at $2.18\times$.}
\label{tab:asym_path}
\setlength{\tabcolsep}{4pt}
\begin{tabular}{l|ccc|c}
\toprule
\textbf{Variant} &
\textbf{AUC@30} $\uparrow$ &
\textbf{Time} $\downarrow$ &
\textbf{Speed} $\uparrow$ &
\textbf{Status} \\
\midrule
VGGT$^\ast$ (baseline)         & 0.9162          & 556.1s          & $1.00\times$         & ref. \\
Step~1: skip deep global & 0.6330          & 169.4s          & $3.28\times$         & FAIL \\
Step~2: deep-band asym.  & 0.7965          & 179.2s          & $3.10\times$         & PARTIAL \\
\rowcolor[rgb]{.92,.96,.99}\textbf{Step~3: selectively protected \emph{(ours)}} & \textbf{0.9185} & \textbf{254.6s} & $\mathbf{2.18\times}$ & \textbf{PASS} \\
\bottomrule
\end{tabular}
\par\smallskip
\begin{minipage}{\linewidth}
\small
\textit{Attention configurations.} The baseline leaves every global block unchanged. Step~1 skips L19--L24; Step~2 retains only camera/register queries in L19--L24, with patch queries bypassing these global blocks. Both leave L1--L18 unchanged. Step~3 retains all queries at every layer and uses phase-shifted K/V sampling with the default per-band rates $(\sigma_s,\sigma_m,\sigma_d){=}(1.5,1.3,1.7)$; camera/register tokens, frame~0, and the $\Psi$-top-$\alpha$ patches are retained at full density.
\end{minipage}
\end{table}

\section{Complexity Derivation}
\label{app:complexity}

In the original VGGT, every layer runs both frame and global attention. With 24 frame blocks and 24 global blocks (alternating schedule), the total cost is $\mathcal{O}_{\text{VGGT}} = 24 \cdot O(SL^2) + 24 \cdot O(S^2L^2)$, dominated by the $O(S^2L^2)$ term for large $S$.

\paragraph{Token merging.}
At each global block in band $b$, merging away $r_b$ eligible patch tokens leaves $L_b'=T-r_b+5$ tokens per non-reference frame, including the protected tokens. The ratio $\rho_b$ controls the eligible source set, so it does not directly specify the fraction removed from all $T$ patches. With the reference frame kept at full length, the total merged sequence length is $M_b=L+(S-1)L_b'$, giving a global-attention cost of $O(M_b^2)$. For large $S$, ignoring the single reference frame yields the leading approximation $O\!\bigl(S^2 L_b^{\prime\,2}\bigr)$. The bands contain 10 shallow, 8 middle, and 6 deep layers; DPT taps read the full-resolution features after unmerging. Token merging alone gives a $2.78\times$ mean end-to-end speedup over VGGT$^\ast$ on Tanks~\&~Temples (Table~\ref{tab:ablation}).

\paragraph{K/V downsampling.}
The query set is unchanged; per band, the nominal grid retains approximately a $1/\sigma_b^2$ fraction of each non-anchor frame's patch K/V, with the full anchor frame and the protected-token union retained in addition. Ignoring discretization and protected-token corrections, the per-block global-attention cost is $O\!\bigl(S^2 L^2 / \sigma_b^2\bigr) + O\!\bigl(S L^2\bigr)$ for the anchor contribution. At a uniform $\sigma_b{=}1.5$ across bands plus the frame-$0$ anchor (the configuration used for the anchor ablation in Table~\ref{tab:anchor_ablation}), the approximate $49\%$ K/V retention is specific to this configuration; the reported end-to-end speedup is $2.09\times$ on Tanks~\&~Temples (Table~\ref{tab:anchor_ablation}). With the default per-band rates $(1.5,1.3,1.7)$, K/V downsampling alone gives $2.18\times$ (Table~\ref{tab:ablation}).

\paragraph{Both axes (\methodname{}).}
The two compressions compose within each global block. Evaluating the phase-shifted $K/V$ index set on the merged token grid gives the leading per-block approximation $O\!\bigl(S^2 L_b^{\prime\,2} / \sigma_b^2\bigr)$ in band $b$, with additional reference-frame and protected-token contributions. Summing this cost over the three bands gives the stack's global-attention cost. The canonical token-merge ratios $\rho_s{=}0.99,\ \rho_m{=}0.5,\ \rho_d{=}0.99$ and K/V rates $\sigma_s{=}1.5,\ \sigma_m{=}1.3,\ \sigma_d{=}1.7$ combine reductions in query/key count and K/V density. The resulting measured end-to-end speedup over VGGT$^\ast$ is $6.7\times$ on 1000-frame ScanNet-50 (Table~\ref{tab:scannet}); wall-clock gains also depend on the remaining model computation and operator overhead.

\section{Comparison with AVGGT}
\label{app:avggt}

AVGGT~\cite{sun2025avggt} is the closest concurrent method to ours: it 
reports a similar early/middle/late division of global-layer roles, converts early global 
layers to frame attention ($t_\text{early}{=}9$), whereas our shallow band L1--L10 keeps global attention under compression, 
and accelerates the remaining layers via K/V subsampling on a uniform 
spatial grid. We discuss the design choices and theoretical attention costs here. The two papers report results with different hardware and baselines, so their separately reported measurements do not establish a numerical ranking under a common protocol.

\paragraph{Theoretical compression comparison.}
Both methods reduce the global-attention cost $O(S^2 L^2)$. AVGGT's subsampling factor is an area factor: $\sigma{=}4$ retains approximately one quarter of patch K/V per frame, and $\sigma{=}9$ retains approximately one ninth, in addition to its early G2F conversion. RegimeVGGT combines token merging with band-specific K/V downsampling, giving the leading cost $O(S^2 L_b^{\prime\,2}/\sigma_b^2)$ under the approximations in Appendix~\ref{app:complexity}. The token-count axis reduces both queries and keys, while the K/V axis further reduces the context read by each query. This composition distributes compression across two complementary dimensions while preserving salient patches, the reference frame, and the cross-frame pose path.

\paragraph{Coverage of evaluation scenarios.}
AVGGT evaluates pose on sequences of up to 800 frames (7\,Scenes) but does not report 1000-frame ScanNet-50 pose. 
Our results in Table~\ref{tab:pose_scannet} show 
that preserving the cross-frame pose path remains effective at this scale: RegimeVGGT 
achieves the best ATE (0.092) among all methods at 1000 frames, 
substantially improving over VGGT$^\ast$ (0.117) and FastVGGT (0.111). 
In addition to the special-token and reference-frame protection shared with AVGGT~\cite{sun2025avggt}, our design combines reversible token merging, band-specific K/V rates, phase-shifted sampling, and saliency protection. This composition targets cross-frame spatial coverage and pose stability while keeping the pretrained weights frozen.

\section{More Ablations}
\label{app:kv_rate_sensitivity}
\paragraph{Ablation on K/V subsampling rate.}
Table~\ref{tab:kv_sensitivity} varies one band's K/V rate at a time on all six Tanks~\&~Temples scenes, fixing the merging schedule and all other settings. Relative to the default, the tested shallow and deep rates change AUC@30 by at most 0.018 and 0.057 percentage points, respectively. Increasing the middle rate from 1.3 to 1.7 reduces total time from 111.00 to 103.00\,s with a 0.242-point AUC loss. Removing middle-band K/V downsampling ($\sigma_m=1.0$) takes 133.00\,s and yields 90.774\% AUC. The middle rate therefore offers the clearest accuracy--latency trade-off within this sweep.

\paragraph{Layer-aware allocation.}
Table~\ref{tab:merge_allocation} compares merging allocations with the K/V policy and other settings fixed. Preserving the middle band improves both accuracy and latency over uniform-low and reversed allocations. The default U-shaped allocation gains 0.304 percentage points of AUC@30 over uniform-high, supporting token preservation in the alignment-critical middle band.

\begin{table}[htbp]
\centering
\small
\caption{K/V-rate sensitivity on all six Tanks~\&~Temples scenes. Each non-default row changes only the indicated rate; merging ratios remain $(0.99,0.50,0.99)$. Larger $\sigma_b$ retains fewer patch K/V tokens. AUC@30 is the scene mean in percent; time is summed over the six scenes.}
\label{tab:kv_sensitivity}
\setlength{\tabcolsep}{6pt}
\begin{tabular}{lccc}
\toprule
\textbf{Varied rate} & $(\sigma_s,\sigma_m,\sigma_d)$ & \textbf{AUC@30 (\%)}$\uparrow$ & \textbf{Total time (s)}$\downarrow$ \\
\midrule
\rowcolor[rgb]{0.92,0.96,0.99}Default & $(1.5,1.3,1.7)$ & 90.883 & 111.00 \\
\midrule
$\sigma_s=1.3$ & $(1.3,1.3,1.7)$ & 90.901 & 112.87 \\
$\sigma_s=1.7$ & $(1.7,1.3,1.7)$ & 90.900 & 109.39 \\
\midrule
$\sigma_m=1.0$ & $(1.5,1.0,1.7)$ & 90.774 & 133.00 \\
$\sigma_m=1.5$ & $(1.5,1.5,1.7)$ & 90.779 & 108.08 \\
$\sigma_m=1.7$ & $(1.5,1.7,1.7)$ & 90.641 & 103.00 \\
\midrule
$\sigma_d=1.3$ & $(1.5,1.3,1.3)$ & 90.940 & 112.96 \\
$\sigma_d=1.5$ & $(1.5,1.3,1.5)$ & 90.867 & 111.86 \\
$\sigma_d=2.0$ & $(1.5,1.3,2.0)$ & 90.921 & 109.87 \\
\bottomrule
\end{tabular}
\end{table}

\begin{table}[htbp]
\centering
\small
\caption{Merging allocations on six Tanks~\&~Temples scenes. H ($\rho=0.99$) and L ($\rho=0.50$) denote stronger and lighter merging. K/V rates remain $(1.5,1.3,1.7)$. AUC@30 is the scene mean; time is summed over scenes.}
\label{tab:merge_allocation}
\setlength{\tabcolsep}{6pt}
\begin{tabular}{lccc}
\toprule
\textbf{Allocation} & \textbf{Shallow / middle / deep} & \textbf{AUC@30 (\%)}$\uparrow$ & \textbf{Total time (s)}$\downarrow$ \\
\midrule
\rowcolor[rgb]{0.92,0.96,0.99}U-shaped (ours) & H / L / H & \textbf{90.883} & 111.0 \\
Uniform-low & L / L / L & 90.764 & 143.8 \\
Uniform-high & H / H / H & 90.579 & \textbf{92.5} \\
Reversed & L / H / L & 90.592 & 126.1 \\
\bottomrule
\end{tabular}
\end{table}

\section{Visualizing Deep-Band Updates}
\label{app:scannet_visualizations}
\renewcommand{\theHfigure}{scannet.\arabic{figure}}

\begin{figure}[H]
    \centering
    \includegraphics[width=\linewidth]{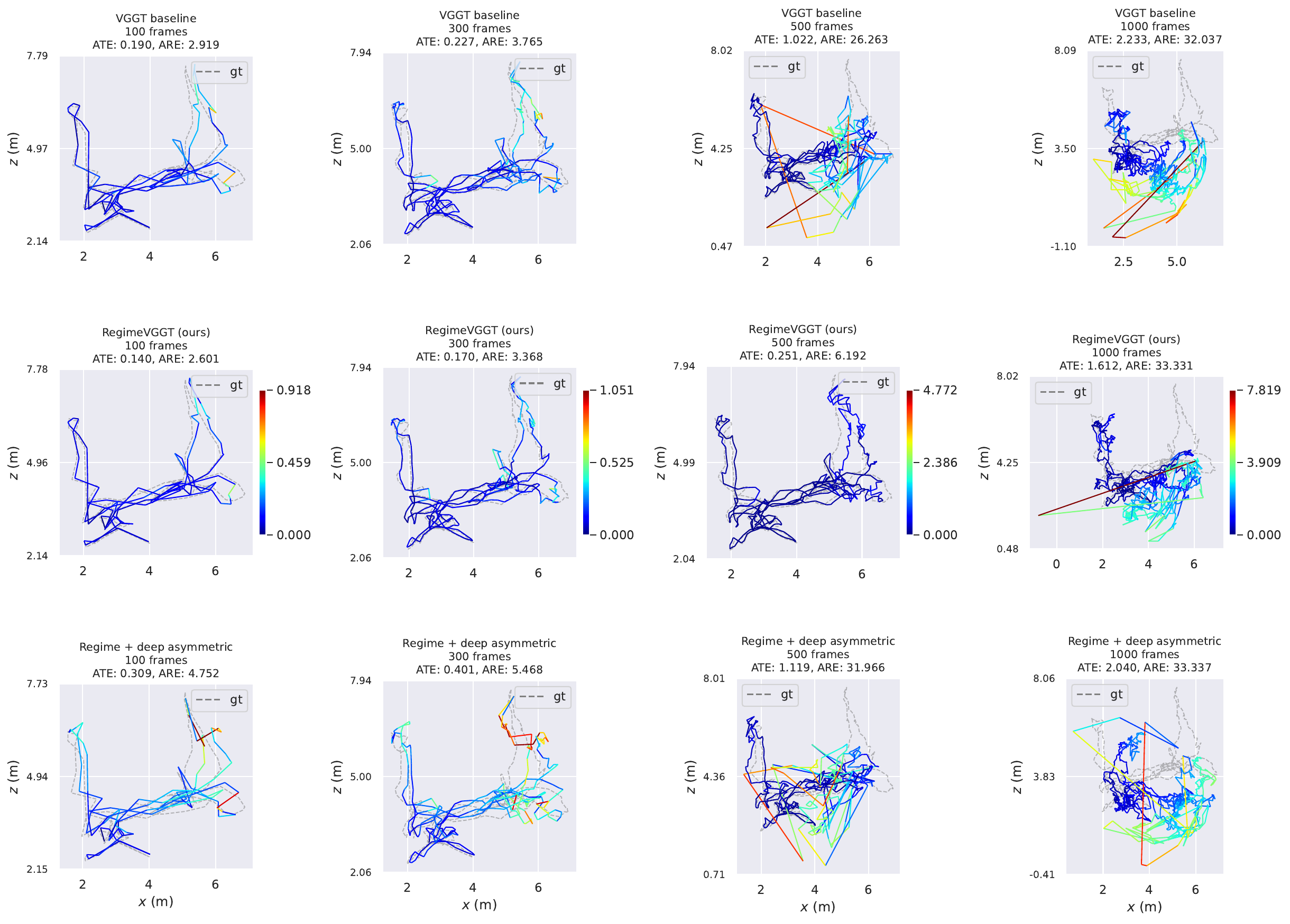}
    \caption{\textbf{Camera-pose degradation under deep-band asymmetric updates.} Rows show VGGT$^\ast$, \methodname{}, and \methodname{} with deep-asymmetric updates; columns show 100, 300, 500, and 1000 input frames. Gray dashed curves denote ground truth and colored curves denote predicted trajectories in the $x$--$z$ plane. Each panel reports ATE and ARE. These trajectory results are from the original case study, evaluated separately from the point-cloud reruns below.}
    \label{fig:scannet_trajectory_scale}
\end{figure}

\paragraph{Camera trajectories.}
Figure~\ref{fig:scannet_trajectory_scale} makes the camera-pose degradation visible in the original case study. At 500 frames, deep-asymmetric updates increase ATE from 0.251 to 1.119\,m and ARE from 6.192$^\circ$ to 31.966$^\circ$ relative to \methodname{}. Retaining camera/register queries alone therefore does not guarantee accurate poses when patch updates are frozen in the deep global blocks. At 1000 frames, all three methods exhibit substantial drift, including \methodname{}, showing that the protected design also has limits on long inputs.

\begin{figure}[H]
    \centering
    \includegraphics[page=1,width=\linewidth]{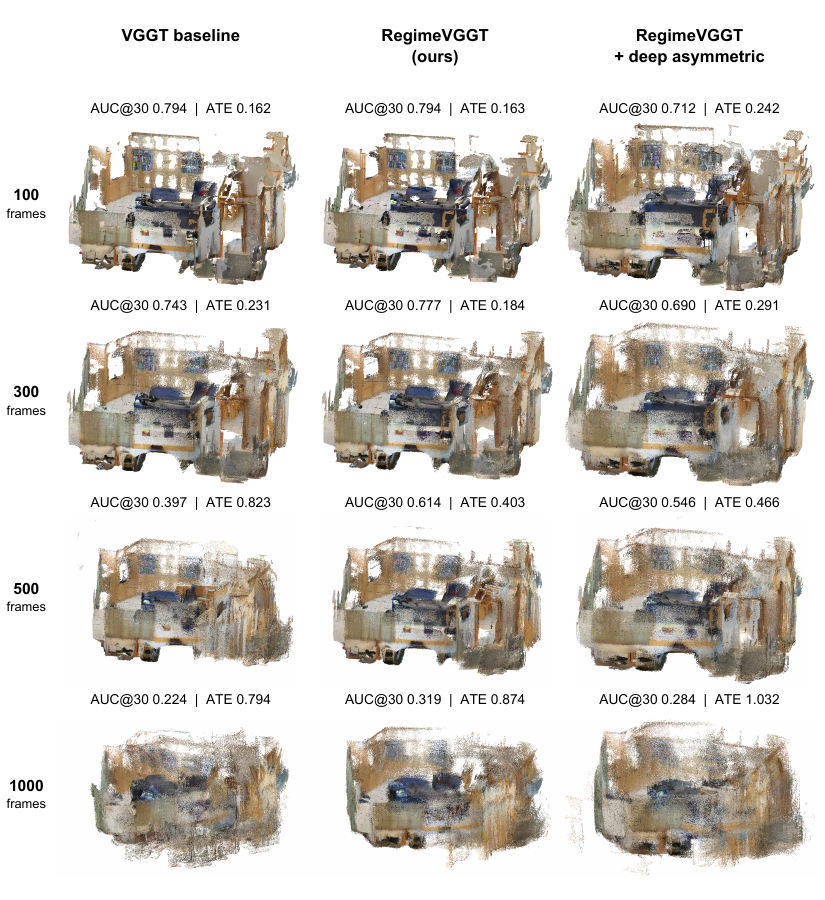}
    \caption{\textbf{High-resolution point-cloud reruns across input lengths.} Rows show 100, 300, 500, and 1000 input frames; columns show VGGT$^\ast$, \methodname{}, and the deep-asymmetric variant. Each panel reports AUC@30 and ATE (meters) from its new inference run. Within each row, display bounds use the union of the methods' 0.2--99.8 percentile coordinate ranges with a 3\% margin. Point colors come from the input images. These one-scene reruns are independent of Figure~\ref{fig:scannet_trajectory_scale}.}
    \label{fig:scannet_reconstruction_scale}
\end{figure}

\paragraph{Point-cloud reconstructions.}
Figure~\ref{fig:scannet_reconstruction_scale} shows new matched reruns using the same checkpoint, $518\times392$ inputs, and bfloat16 inference. Geometry uses predicted depths and poses. A similarity transform fitted to ground-truth camera centers aligns each prediction for ATE, ARE, and display; AUC@30 uses all unordered relative-pose pairs. For display, all methods retain the upper half of per-frame confidence values and sample uniformly within each frame, up to two million points per reconstruction. Within each input length, viewpoint, bounds, point size, and rendering resolution are shared. This filtering does not affect pose evaluation.

\begin{figure}[H]
    \centering
    \includegraphics[page=2,width=\linewidth]{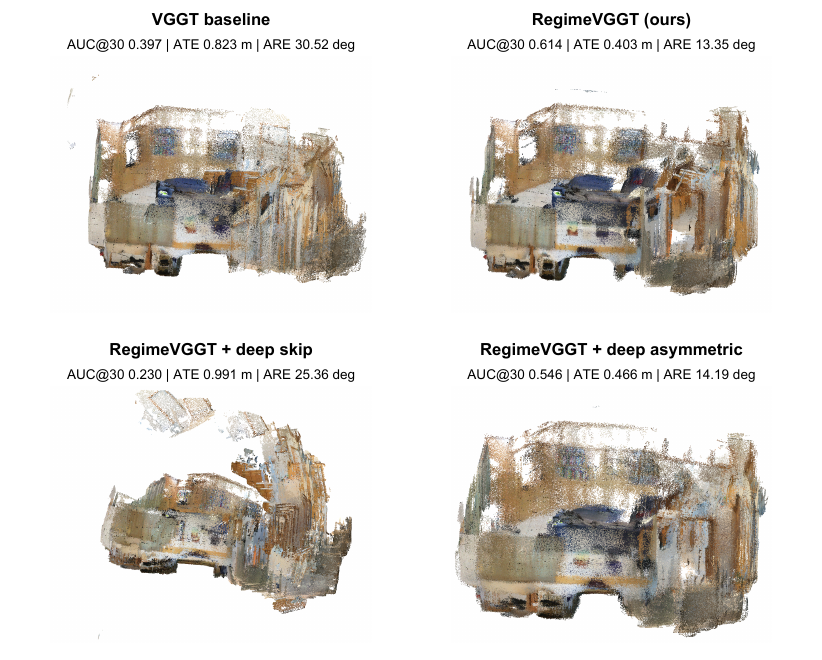}
    \caption{\textbf{Matched 500-frame deep-band controls.} All four methods share the input frames, checkpoint, and rendering settings. The three \methodname{} variants share the shallow/middle policy; VGGT$^\ast$ uses full attention. Deep skip makes all six global blocks L19--L24 no-ops, while the asymmetric variant updates only camera/register queries in these blocks. The annotations report pose metrics from the same runs as the displayed reconstructions. This comparison isolates the interventions on top of the final \methodname{} policy.}
    \label{fig:scannet_deep_controls}
\end{figure}

\clearpage
\paragraph{Deep skip versus asymmetric queries at 500 frames.}
Figure~\ref{fig:scannet_deep_controls} adds the complete deep-skip intervention to the same 500-frame rerun. Deep skip makes all six global blocks L19--L24 no-ops; the frame-attention path and L1--L18 remain active. Relative to \methodname{}, deep skip increases ATE from 0.403 to 0.991\,m and ARE from 13.35$^\circ$ to 25.36$^\circ$. Its aggregate point cloud is also visibly fragmented in this case. Restoring camera/register queries through deep-asymmetric updates recovers much of this loss, reaching 0.466\,m and 14.19$^\circ$, but does not fully match \methodname{} in this case. This control separates the cost of removing all deep global updates from the smaller residual pose gap when camera/register queries remain active.

\begin{figure}[t]                                                                         
    \centering                                                      
    \includegraphics[width=\linewidth]{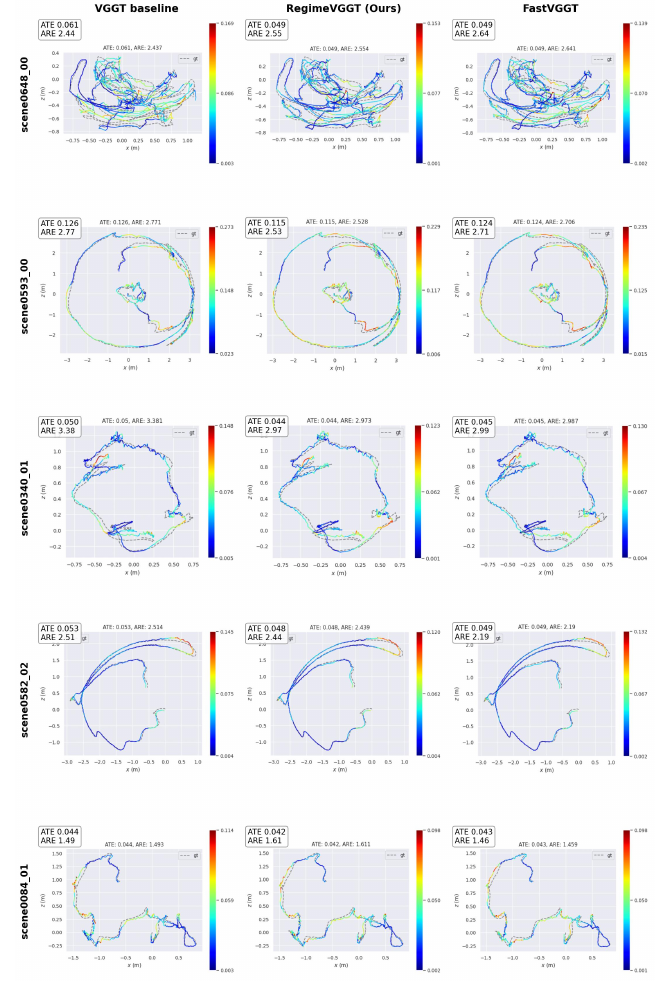}                         
    \caption{Additional visualizations of pose estimation results on the ScanNet dataset.}
    \label{fig:qual_scannet_trajectories}
  \end{figure}

\begin{figure}[t]                                                                         
    \centering                                                      
    \includegraphics[width=\linewidth]{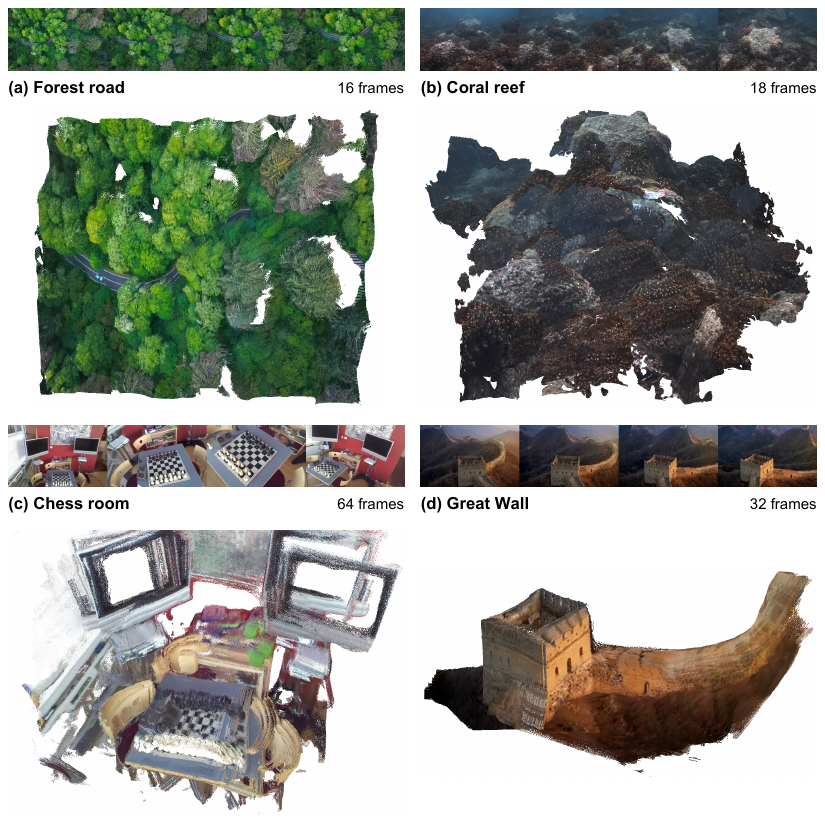}                         
    \caption{Additional visualizations of point cloud results.}
    \label{fig:qual_multiscene_reconstruction}
  \end{figure}

\end{document}